\pdfoutput=1

\documentclass[11pt]{article}

\usepackage{acl}

\usepackage{times}
\usepackage{latexsym}

\usepackage[T1]{fontenc}
\usepackage[most]{tcolorbox}
\tcbuselibrary{breakable,skins}

\usepackage[utf8]{inputenc}

\usepackage{microtype}

\usepackage{graphicx}

\title{A Short Survey on Small Reasoning Models: Training, Inference, Applications and Research Directions}

\author{Chengyu Wang$^1$\thanks{\ \ \ C. Wang and T. Zhang contributed equally to this work. Correspondence to C. Wang.}, Taolin Zhang$^2$\footnotemark[1], Richang Hong$^2$, Jun Huang$^1$\\
  $^1$ Alibaba Cloud Computing\\
  $^2$ School of Computer Science and Information Engineering, Hefei University of Technology\\
  \texttt{chengyu.wcy@alibaba-inc.com}\\}

\begin{document}
\maketitle
\begin{abstract}
Recently, the reasoning capabilities of large reasoning models (LRMs), such as DeepSeek-R1, have seen significant advancements through the slow thinking process. Despite these achievements, the substantial computational demands of LRMs present considerable challenges. In contrast, small reasoning models (SRMs), often distilled from larger ones, offer greater efficiency and can exhibit distinct capabilities and cognitive trajectories compared to LRMs. This work surveys around 170 recently published papers on SRMs for tackling various complex reasoning tasks. We review the current landscape of SRMs and analyze diverse training and inference techniques related to SRMs. Furthermore, we provide a comprehensive review of SRMs for domain-specific applications and discuss possible future research directions. This survey serves as an essential reference for researchers to leverage or develop SRMs for advanced reasoning functionalities with high efficiency.
\end{abstract}

\section{Introduction}

\begin{small}
\begin{quote}
\textit{Be faithful in small things because it is in them that your strength lies.\hfill---Mother Teresa  
}
\end{quote}
\end{small}
Recently, NLP has been extensively transformed by large language models (LLMs)~\cite{DBLP:journals/corr/abs-2303-18223}, which demonstrate remarkable capabilities across a wide range of downstream tasks. Noteworthy among these are large reasoning models (LRMs)~\cite{DBLP:journals/corr/abs-2501-09686}, such as DeepSeek-R1~\cite{DBLP:journals/corr/abs-2501-12948} and QwQ-32B\footnote{\url{https://qwenlm.github.io/blog/qwq-32b/}}, which are specialized in solving reasoning problems (such as mathematical problems, code generation, logical reasoning, etc.) through the utilization of slow thinking processes. However, the prowess of these models comes at a huge cost, requiring significant computational resources, either for training or inference. For instance, the DeepSeek-R1 model comprises 671B parameters and requires a sever with eight A100 (80GB) GPUs at least or  better hardware configurations for online serving.

\begin{figure}[!t]
    \centering
    \includegraphics[width=0.275\textwidth]{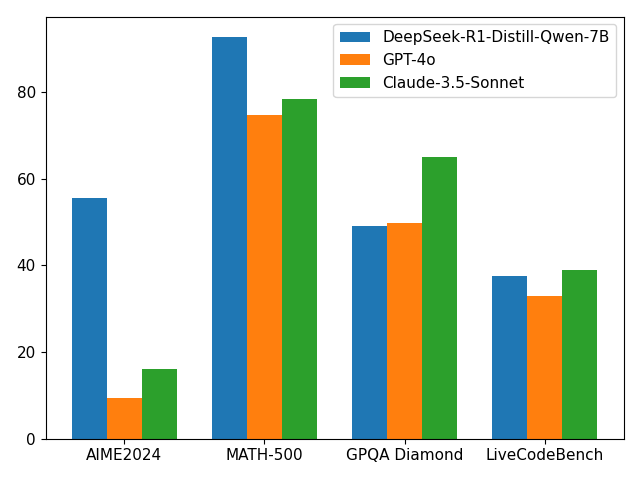}
    \vspace{-.5em}
    \caption{A simple comparison between representative LLMs and SRMs on various reasoning benchmarks.}
    \vspace{-1.25em}
    \label{fig:intro}
\end{figure}

Consequently, there is growing interest within the research community in leveraging much smaller models~\cite{DBLP:conf/icml/FuPOSK23,DBLP:conf/acl/MagisterMAMS23,DBLP:conf/acl/ShridharSS23,DBLP:journals/expert/ZhangLP25} to tackle complex reasoning tasks, seeking more efficient yet effective alternatives. Following the release of DeepSeek-R1, the open-source community has observed numerous achievements demonstrating that small reasoning models (SRMs), equipped with slow-thinking capabilities (i.e., long chain-of-thought processes~\cite{DBLP:conf/nips/Wei0SBIXCLZ22}), can outperform much larger LLMs in some reasoning tasks, as illustrated in Figure~\ref{fig:intro}.\footnote{There is no official definition for the size of SRMs. In this work, we define SRMs as strong reasoning language models typically with fewer than 10 billion parameters.}
However, since SRMs often exhibit distinct capabilities and cognitive trajectories compared to LRMs~\cite{DBLP:conf/emnlp/Yan0ZHHZ23,DBLP:conf/iclr/0006LCF24,DBLP:conf/aaai/HuHWZSN24}, the training and inference methodologies for these models can be fundamentally different. Consequently, considerable effort has been devoted to developing strong SRMs that are comparable to, or even surpass LRMs.
We have observed that in the literature there are several surveys that focus on the reasoning tasks of LLMs~\cite{DBLP:journals/corr/abs-2407-11511,DBLP:journals/corr/abs-2501-09686,DBLP:conf/acl/0009C23,DBLP:conf/emnlp/GiadikiaroglouL24,DBLP:conf/eacl/AhnVLLZY24}. Yet, the surveys and reviews focusing specifically on SRMs remain insufficient.

\begin{figure}[!t]
    \centering
    \includegraphics[width=.475\textwidth]{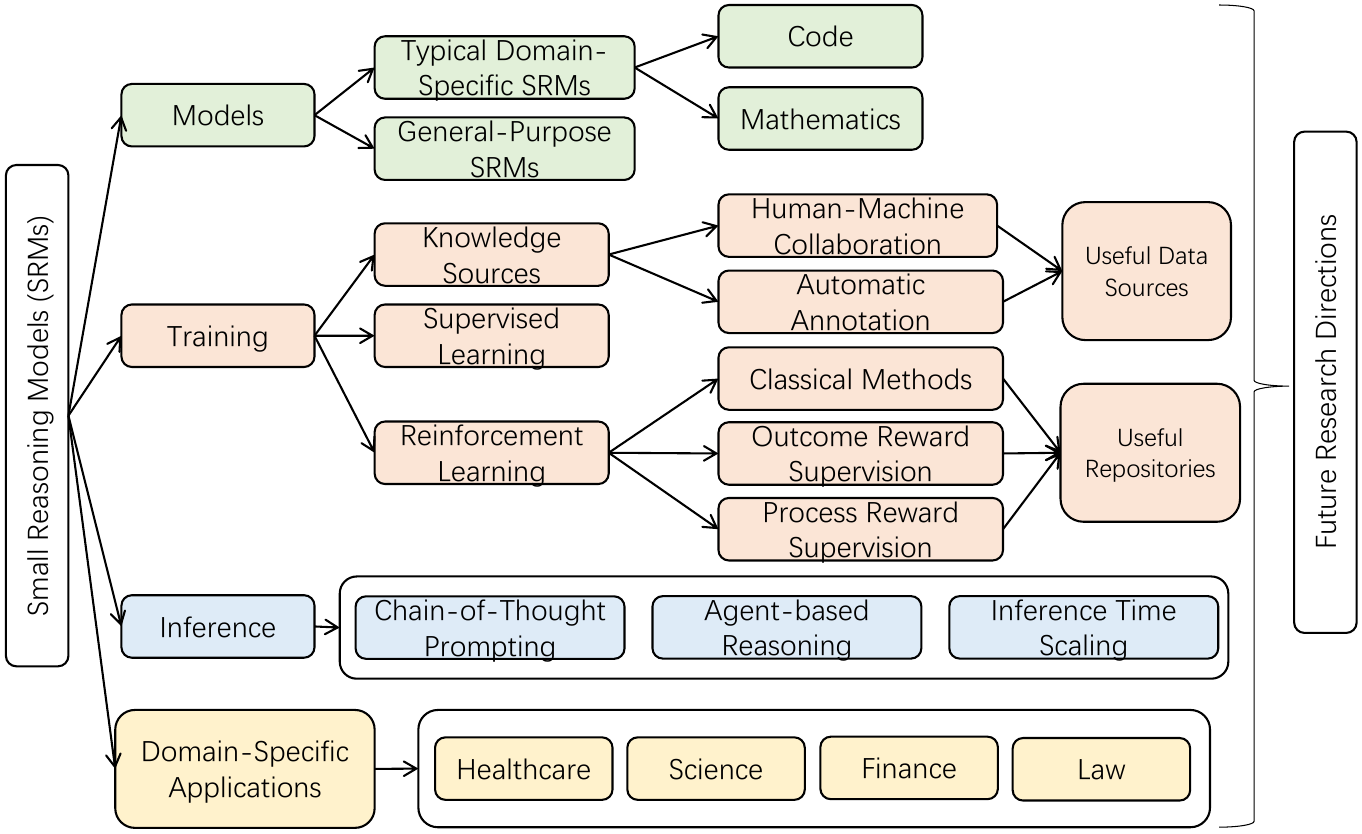}
    \vspace{-.5em}
    \caption{The roadmap of this survey.}
    \vspace{-1.25em}
    \label{fig:roadmap}
\end{figure}

In this paper, we offer a short yet comprehensive survey of SRMs. By reviewing around 170 research papers, mostly published or made public in the last three years, our survey aims to consolidate knowledge on techniques, applications, and possible future research directions of SRMs. The roadmap of this survey is shown in Figure~\ref{fig:roadmap}.

\noindent\underline{\textbf{What is covered in this survey?}}
We first provide a quick glance at popular SRMs in the open-source community. Next, we explore various training and inference techniques aimed at enhancing the reasoning abilities of pre-trained models. Additionally, we survey domain-specific applications based on these models, discuss possible future research directions and propose our suggestions.

\noindent\underline{\textbf{What is NOT covered in this survey?}}
This survey does not cover the general model architecture design and algorithms for LLMs as a whole, nor does it delve into tasks unrelated to complex reasoning tasks. Additionally, we do not explore compression techniques (such as pruning and quantization) and large-scale pre-training techniques for obtaining smaller models, focusing instead on specific techniques for reasoning.

In short, the exploration of SRMs represents a significant and timely research topic for the NLP community. By embracing the efficiency and capabilities of SRMs, researchers can drive forward the development of models that are high-performing and sustainable for real-world applications.

\section{A Quick Glance at SRMs}

Following the release of OpenAI's o1\footnote{\url{https://openai.com/o1/}}, the AI community has witnessed a paradigm shift towards developing models with strong reasoning abilities. We review popular SRMs available in the open-source community that serve as very useful backbones for researchers towards further exploration.

\subsection{Typical Domain-Specific SRMs}

Prior to OpenAI's o1, task-specific SRMs were introduced, particularly for code-related tasks such as code completion and natural language to code translation, due to their wide applications. More recently, the Qwen2.5-Coder~\cite{DBLP:journals/corr/abs-2409-12186} series includes LRMs and SRMs of varying sizes, from 1.5B to 32B. DeepSeek-Coder~\cite{DBLP:journals/corr/abs-2401-14196}, StarCoder2~\cite{DBLP:journals/corr/abs-2402-19173} and OlympicCoder\footnote{\url{https://huggingface.co/open-r1/OlympicCoder-7B}} are other notable series, featuring SRMs with robust reasoning abilities for code-related tasks. For additional information on code-related SRMs, readers are referred to \citet{DBLP:journals/corr/abs-2406-00515}. Mathematics presents another intriguing domain for SRMs, involving intricate reasoning steps to solve mathematical problems. Prominent open-source series include Qwen2.5-Math~\cite{DBLP:journals/corr/abs-2409-12122}, DeepSeek-Math~\cite{DBLP:journals/corr/abs-2402-03300}, and InternLM-Math~\cite{DBLP:journals/corr/abs-2402-06332}. For SRMs in other domains (such as healthcare, science, law and finance), we refer readers to Section~\ref{sect:app} for details.

\subsection{General-Purpose SRMs}
As powerful LRMs, such as DeepSeek-R1 and QwQ-32B, become publicly available with explicit long CoT trajectories as part of their outputs, general-purpose SRMs capable of solving various types of reasoning tasks have also been released, most of which are trained using the knowledge distillation technique from LRMs. Some examples of these models include distilled models based on LLaMA and Qwen series released by the DeepSeek AI team~\cite{DBLP:journals/corr/abs-2501-12948} (such as DeepSeek-R1-Distill-LLaMA-8B and DeepSeek-R1-Distill-Qwen-7B), s1~\cite{DBLP:journals/corr/abs-2501-19393}, OpenThinker\footnote{\url{https://huggingface.co/open-thoughts/OpenThinker-7B}}, Bespoke-Stratos\footnote{\url{https://huggingface.co/bespokelabs/Bespoke-Stratos-7B}}, LLaMA-O1\footnote{\url{https://github.com/SimpleBerry/LLaMA-O1}}, Marco-o1~\cite{DBLP:journals/corr/abs-2411-14405}, and many others.
In summary, the availability of SRMs marks a significant advancement in the field. These SRMs enable researchers to conduct efficient and economical research on a variety of backbones. 

\section{Training SRMs without Pain}

In this section, we discuss the training pipelines for producing high-quality SRMs.

\subsection{Obtaining the Knowledge Sources}

Creating high-quality datasets containing reasoning processes is crucial for training SRMs. Although human annotation guarantees superior quality, it is extremely costly and impractical, particularly for annotating the whole long chain-of-thought (CoT) processes~\cite{DBLP:conf/nips/Wei0SBIXCLZ22} required in training SRMs, since annotators must guide or even write each reasoning step~\cite{DBLP:conf/iclr/LightmanKBEBLLS24}. Next, we will briefly discuss human-model collaboration methods and focus more on automatic annotation.\footnote{To support future research, we provide an ever-growing yet incomplete list of popular open-source datasets with long CoT and output annotations generated by LRMs, as shown in Table~\ref{tab:data} (in the appendix). These datasets can be downloaded from HuggingFace Datasets, serving as a starting point to train SRMs for researchers without costly reproduction.}

\noindent\textbf{Human-Model Collaboration.}
In human-model collaboration, LLMs/LRMs perform annotations using a few selected high-quality examples as in-context demonstrations. Human annotators then correct low-quality annotations only, which constitute a small part of the entire dataset~\cite{DBLP:conf/lrec/MikulovaSSSH22,DBLP:conf/eacl/KimMCRZ24,DBLP:conf/emnlp/Li24,DBLP:conf/chi/Wang0RMM24,DBLP:conf/emnlp/MovvaKP24}. This type of approach ensures the data quality while minimizing the need for human involvement.

\noindent\textbf{Automatic Annotation.}
Earlier works using automatic annotation techniques focused primarily on complex tasks without generating long reasoning trajectories, such as tool usage~\cite{DBLP:conf/nips/SchickDDRLHZCS23,DBLP:conf/sigir/0003QLPW24,DBLP:conf/naacl/QiaoGLJC024}. As the reasoning abilities of the annotators improve, advanced LRMs can now generate long reasoning trajectories in a zero-shot manner~\cite{DBLP:journals/ral/KwonPJ24}.
For example, leveraging DeepSeek-R1 for automatic annotation is an efficient solution, as it can easily generate lengthy CoT trajectories with large context windows~\cite{DBLP:journals/corr/abs-2501-12948}. This approach is also known as ``knowledge distillation'', leveraged for training various types of SLMs~\cite{DBLP:conf/acl/HsiehLYNFRKLP23,DBLP:conf/acl/HoSY23,DBLP:conf/acl/LiHYRC023,DBLP:conf/emnlp/YueWHW24,DBLP:journals/corr/abs-2412-04871,DBLP:conf/acl/YangPFWCZL24,DBLP:conf/icml/TangZWW24}. The work~\cite{DBLP:conf/icml/HavrillaRNDZHR24} shows that SRMs can be trained effectively with synthetic datasets only.
Another research trend is the incorporation of specialized agents to annotate training data based on various operations including planning, tool use, reflection and refinement. In this scenario, LRMs not only annotate the dataset but also document how these agents make correct decisions through interaction~\cite{DBLP:conf/acl/Qiao0FLZJLC24,DBLP:journals/corr/abs-2407-04078,DBLP:journals/corr/abs-2403-02502,DBLP:journals/corr/abs-2412-01928,DBLP:journals/corr/abs-2409-12917}. These interaction steps can be treated as reasoning processes.

In addition, since long reasoning trajectories inherently include multiple steps, determining the correctness of each step provides fine-grained insights for SRMs to learn. Although these annotation tasks are more demanding, they are beneficial for training process reward models (PRMs)~\cite{DBLP:journals/corr/abs-2501-07301}. 
In the literature, an early work~\cite{DBLP:journals/corr/abs-2308-09583} annotates the correctness of each step in mathematical problem-solving. Subsequent works~\cite{DBLP:conf/acl/WangLSXDLCWS24,DBLP:conf/emnlp/WangLWLH0S24} employ Monte Carlo sampling to assess intermediate reasoning steps based on the average outcome of inference results from these steps. Further extensions rely on Monte Carlo Tree Search (MCTS) and its variants, leveraging tree search to enhance the inference quality~\cite{DBLP:conf/nips/ZhangZHYD024,DBLP:conf/acl/ChenWMP0024}.
Thus, we suggest that inference-time scaling approaches can be leveraged for dataset curation.

\begin{figure*}[!t]
\vspace{-1em}
\begin{small}
\begin{tcolorbox}[colback=blue!10,colframe=blue!40!black,title=\textbf{\texttt{Takeways (Training)}}]
Training SRMs can be streamlined through a combination of data annotation and advanced training algorithms. Leveraging automatic annotation and knowledge distillation techniques is increasingly important with the release of strong models. CoT-based SFT is a good starting point to train SRMs based on the collected and annotated datasets. Advanced RL methods offer effective training by evaluating intermediate reasoning steps and reducing computational needs. There is no widely accepted consensus on which RL algorithm performs the best for SRMs, while the successful story of DeepSeek-R1 favors outcome reward supervision~\cite{DBLP:journals/corr/abs-2501-12948}, treating PRMs and MCTS as ``unsuccessful attempts''. We suggest that it might be too early to draw the conclusion especially for SRMs. Researchers can accelerate the research by utilizing datasets and RL repositories that facilitate SRM development.
\end{tcolorbox}
\end{small}
\vspace{-1.5em}
\end{figure*}

\subsection{Supervised Learning}

Supervised Fine-Tuning (SFT) is a standard supervised learning technique that aligns models with task-specific instructions. With the availability of high-quality reasoning datasets, it is natural to extend SFT to CoT-based SFT. In this paradigm, SRMs are tasked with explicitly generating intermediate reasoning steps and ultimately providing outputs for input instructions, thereby enhancing their reasoning abilities to tackle complex tasks. The report~\cite{DBLP:journals/corr/abs-2501-12948} indicates that CoT-based SFT approach combined with knowledge distillation produces strong SRMs.
However, SFT is limited by its reliance on high-quality labeled datasets, which are costly and time-consuming to produce. For example, the report shows that the dataset size for training distilled DeepSeek-R1 models is 800K.
In addition, fine-tuning SRMs remains more computationally expensive than general language models, especially when the training sequence (with CoT trajectories) is long. which necessitates parameter-efficient learning~\cite{DBLP:journals/corr/abs-2311-13126,DBLP:journals/corr/abs-2403-14608}, such as LoRA~\cite{DBLP:conf/iclr/HuSWALWWC22}, QLoRA~\cite{DBLP:conf/nips/DettmersPHZ23}, AdaLoRA~\cite{DBLP:conf/iclr/ZhangCBH0CZ23}, etc.
Another pitiful of SFT is that it forces the SRMs to ``exploit'' the training dataset but not ``explore'' more possibilities, which motivates more studies on reinforcement learning.

\subsection{Reinforcement Learning}

Reinforcement learning (RL) offers a powerful alternative for training SRMs. It allows models to learn optimal strategies through trials and errors and improves generalization abilities, rather than only following golden responses during SFT.

\noindent\textbf{Starting with Classical Methods.}
\citet{DBLP:conf/nips/Ouyang0JAWMZASR22} employ reinforcement learning from human feedback (RLHF) with proximal policy optimization (PPO)~\cite{DBLP:journals/corr/SchulmanWDRK17} to align LLMs with human intents. To reduce the efforts of manual labeling, the AI feedback reinforcement learning paradigm (RLAIF)~\cite{DBLP:journals/corr/abs-2212-08073} is introduced to leverage model-generated labels for training reward models. These works establish the fundamentals for reinforcement training on language models, which can be applicable to SRMs without any modification. 
To bypass the complex process of optimizing reward models, direct preference optimization (DPO)~\cite{DBLP:conf/nips/RafailovSMMEF23} aligns models using a simple margin-based loss.
Extensions of DPO, such as KTO~\cite{DBLP:journals/corr/abs-2402-01306}, ODPO~\cite{DBLP:conf/acl/AminiVC24}, and SimPO~\cite{DBLP:conf/nips/0001X024}, are also applicable to SRMs due to their auto-regressive nature. However, they are not specifically optimized for the lengthy CoT processes that SRMs generate.

\noindent\textbf{Enhancing Multi-step Reasoning with Outcome Reward Supervision.}
A straightforward approach is to leverage the final outcomes as supervision signals to compute the rewards for RL, without the consideration of intermediate reasoning steps. In the literature, Reinforced fine-tuning (ReFT)~\cite{DBLP:conf/acl/TrungZJSJL24} is an efficient training algorithm for reasoning tasks. It first warms up the underlying model with standard SFT, then further fine-tunes it using an on-line RL process with the PPO algorithm. Abundant reasoning paths are automatically sampled based on the question, with rewards derived from ground-truth answers. VinePPO~\cite{DBLP:journals/corr/abs-2410-01679} identifies bias in the value network of PPO for determining the values of intermediate reasoning steps and instead leverages Monte Carlo sampling for unbiased value function estimates. Critical Plan Step Learning (CPL)~\cite{DBLP:journals/corr/abs-2409-08642} employs MCTS to explore various planning steps in multi-step reasoning tasks, iteratively training policy and value models based on intermediate-step value evaluations to enhance the reasoning performance. Furthermore, Group Relative Policy Optimization (GRPO)~\cite{DBLP:journals/corr/abs-2402-03300} is a variant of the PPO algorithm that eliminates the critic model. It estimates rewards from group scores of multiple inference results from the same input prompt, significantly reducing training resources. GRPO is successfully applied to training strong SRMs such as DeepSeekMath-7B and also to training ultra-large LRMs such as DeepSeek-R1~\cite{DBLP:journals/corr/abs-2501-12948}.\footnote{Note that GRPO can be applied to both outcome supervision and process supervision scenarios.}

\noindent\textbf{Fine-Grained RL with Process Reward Supervision.}
A potential drawback of the above methods is that rewards are provided only for final outcomes. PRMs~\cite{DBLP:journals/corr/abs-2501-07301}, as previously introduced, focus on evaluating intermediate reasoning steps, offering fine-grained feedback throughout the CoT trajectories generated from SRMs. This type of methods is also referred to as process reward supervision. A typical work in this field is Math-Shepherd~\cite{DBLP:conf/acl/WangLSXDLCWS24}, which conducts step-by-step verification and reinforcement based on PRMs.
In addition, Self-Explore~\cite{DBLP:journals/corr/abs-2404-10346} uses PRMs to improve mathematical reasoning by correcting ``first pits'', which are initial errors in problem-solving. It rewards steps that correct these errors, enabling self-supervised fine-tuning without extensive human annotations.
Process Advantage Verifiers (PAVs)~\cite{DBLP:journals/corr/abs-2410-08146} are proposed to evaluate step-level progress to improve the correctness of solution search in RL.
Apart from online RL, off-policy methods derived from DPO~\cite{DBLP:conf/nips/RafailovSMMEF23} have been applied using process reward supervision. For example, SVPO~\cite{DBLP:conf/emnlp/ChenL0024} leverages MCTS to navigate reasoning paths and annotate step-level preferences. Similar works on search-based methods can be found in~\cite{DBLP:journals/corr/abs-2405-00451,DBLP:journals/corr/abs-2406-14283,DBLP:journals/corr/abs-2501-04519}.
We further list some influential public repositories that support RL training for SRMs in Table~\ref{tab:repo} (in the appendix).

\begin{figure*}[!t]
\vspace{-1em}
\begin{small}
\begin{tcolorbox}[colback=blue!10,colframe=blue!40!black,title=\textbf{\texttt{Takeways (Inference)}}]
To boost the reasoning capabilities of SRMs during inference, a diverse array of strategies has been employed.
CoT prompting has been widely used to guide SRMs in generating intermediate reasoning steps, significantly improving their performance on complex reasoning tasks. Agent-based reasoning methods have also shown great potential in enhancing SRMs' reasoning by leveraging interactions between multiple agents. Additionally, inference time scaling techniques, such as self-enhanced tree search and step-wise verifiers, have been proposed to optimize computational resources during inference and achieve better efficiency.
\end{tcolorbox}
\end{small}
\vspace{-1em}
\end{figure*}

\section{Boosting SRM Inference with Scale}
Tackling complex reasoning tasks often requires multi-step computation.
This section explores key approaches for scaling reasoning of SRMs.

\subsection{Chain-of-Thought (CoT) Prompting}
CoT prompting, an extension of few-shot prompting \cite{DBLP:conf/nips/BrownMRSKDNSSAA20}, has demonstrated broader applicability compared to algorithmic and structured reasoning, which initially focused on the generation of intermediate steps~\cite{DBLP:conf/naacl/ChiangC19,DBLP:journals/corr/abs-2112-00114}. 
While simple sampling is computationally straightforward, it is often inefficient and suboptimal, as it randomly allocates the test-time computation budget to less promising branches \cite{DBLP:journals/corr/abs-2408-03314,DBLP:journals/corr/abs-2408-00724}. To tackle this issue, researchers have explored methods that prioritize more promising reasoning paths or intermediate steps, effectively narrowing down the search space~\cite{DBLP:conf/iclr/0002WSLCNCZ23,DBLP:conf/nips/YaoYZS00N23,DBLP:conf/icml/SelAK0024}. CoT-SC \cite{DBLP:conf/iclr/0002WSLCNCZ23} expands on CoT by adopting a tree structure. In this framework, several CoT branches extend from the same initial (root) prompt, and the chain producing the most favorable result is selected as the final answer. SoT \cite{DBLP:conf/iclr/Ning0ZWY024} directs SRMs to generate an answer skeleton and then employs parallel API calls or batched decoding to fill in each skeleton point.
Recently, a considerable number of studies have investigated Tree of Thoughts (ToT) \cite{DBLP:journals/corr/abs-2305-08291,DBLP:conf/nips/YaoYZS00N23} for reasoning enabled by tree search, utilizing tree structure to decompose a question into subquestions and addresses them with distinct prompts.

\subsection{Agent-based Reasoning}
Agent-based reasoning is mainly divided into two categories: agent collaboration to manage different roles and training specific agent graphs.

\noindent\textbf{Agent Collaboration.} Cooperation among multiple agents has emerged as an effective way to enhance the performance of single SRMs \cite{DBLP:conf/icml/Du00TM24,DBLP:conf/emnlp/Liang0JW00Y0T24,DBLP:conf/naacl/WangMW0WJ24}. Existing approaches fall into two main categories: intra-flow and inter-flow communication.
Intra-flow communication examines message exchange between agents within a single conversation round. General communication structures include:
(1) Immediate output: Agents do not communicate with each other and output responses based on their own ability to answer questions \cite{DBLP:conf/acl/ZhangX0LHD24,DBLP:conf/icml/Du00TM24}.
(2) Chain-style connection: Communication between agents is achieved by linking each agent together~\cite{DBLP:conf/acl/QianLLCDL0CSCXL24,DBLP:conf/iclr/HongZCZCWZWYLZR24,DBLP:conf/iclr/HoltLS24}.
(3) Tree-style connection: A supervisory agent (referred to as a root or manager) directs subordinate agents \cite{DBLP:journals/corr/abs-2308-08155,DBLP:journals/corr/abs-2403-17674,DBLP:conf/ijcai/ZhouHZZL24}.
(4) Graph-style connection: Each agent is treated as a node in a graph, and information is transmitted through the edges \cite{DBLP:conf/acl/Jiang0L23,DBLP:conf/icml/ZhugeWKFKS24,agentdropout}.
Inter-flow communication examines the transmission of information across successive utterance rounds. General topologies include:
(1) Full connection: Each agent receives all agents' utterances from the preceding round \cite{DBLP:conf/icml/Du00TM24}.
(2) Partial connection: Certain responses undergo filtering based on scoring or rating systems \cite{DBLP:journals/corr/abs-2304-09797,DBLP:journals/corr/abs-2310-02170}.
(3) Summarization: Prior dialogue is compressed into a summarized form between interaction rounds \cite{DBLP:journals/corr/abs-2303-11366,DBLP:journals/corr/abs-2305-10142,DBLP:conf/naacl/ChenZH24}.

\noindent\textbf{Agent Graphs.} Improving agent cooperation through learning graph connectivity to facilitate communication is a long-standing and effective method. Before the advent of LLMs, numerous endeavors were devoted to exploring optimal communication graph structures for multi-agent systems, utilizing techniques such as graph diffusion \cite{DBLP:journals/ml/PesceM23}, weighted graph neural networks \cite{DBLP:conf/cogsci/LiuD0XL22}, and transformers \cite{DBLP:conf/iclr/Hu00T24}.
In the rising tide of LLM-powered agents, ChatEval \cite{DBLP:conf/iclr/ChanCSYXZF024} and AutoGen \cite{DBLP:journals/corr/abs-2308-08155} implicitly use graph structures to represent simultaneous communication, while STOP \cite{DBLP:journals/corr/abs-2310-02304} and DSPy \cite{DBLP:journals/corr/abs-2310-03714} jointly optimize prompts and inference structures. Moreover, MacNet \cite{DBLP:journals/corr/abs-2406-07155} and GPTSwarm \cite{DBLP:conf/icml/ZhugeWKFKS24} model agent communication using directed acyclic graphs.
Unlike CDC \cite{DBLP:journals/ml/PesceM23} that dynamically modifies the communication graph from a diffusion process viewpoint, and TWG-Q \cite{DBLP:conf/cogsci/LiuD0XL22} which concentrates on temporal weight learning and weighted GCN implementation, CommFormer \cite{DBLP:conf/iclr/Hu00T24} takes a distinct path. It specializes in learning a static graph to enhance communication efficiency before the inference stage, thereby differentiating itself from the conventional techniques used by these previous methods.

\subsection{Inference Time Scaling}
\citet{DBLP:journals/corr/abs-2408-03314}~optimize computational resource allocation during inference to achieve substantial efficiency improvements. Self-enhanced tree search \cite{DBLP:conf/nips/LampleLLRHLEM22,DBLP:journals/corr/abs-2412-09078} consolidates various reasoning paths and applies sparse activation techniques. Additionally, step-wise verifiers are utilized to dynamically prune the search tree \cite{DBLP:conf/mlsys/PopeDCDBHXAD23,DBLP:conf/iclr/LightmanKBEBLLS24}. Moreover, \citet{DBLP:journals/corr/abs-2411-19477} introduce a two-stage elimination method using pairwise comparisons to iteratively refine candidates with query variations. Through iterative refinement, these approaches \cite{DBLP:conf/iclr/WelleckLWBSK023,DBLP:conf/nips/MadaanTGHGW0DPY23,DBLP:conf/iclr/ChenLSZ24,DBLP:journals/corr/abs-2501-19306} enhance improved performance on complex tasks. S1 \cite{DBLP:journals/corr/abs-2501-19393} is proposed as a simple scaling method for test-time applications, employing inference length constraints to maximize resource utilization.

In contrast to repeated sampling, scaling-based approaches enable models to iteratively generate solution attempts, with each subsequent attempt refined based on preceding outcomes \cite{DBLP:journals/corr/abs-2501-11651,DBLP:journals/corr/abs-2501-09891}. Methods like MCTS~\cite{DBLP:conf/emnlp/ChoiF0S23,DBLP:conf/iclr/ZhangCSDTG23,DBLP:conf/icml/ZhouYSWW24} and guided beam search \cite{DBLP:conf/nips/XieKZZKHX23} bridge sequential and parallel scaling \cite{DBLP:journals/corr/abs-2501-19393} through tree-based search \cite{DBLP:journals/corr/abs-2404-03683,DBLP:journals/corr/abs-2408-00724}. Through its innovative process reward model, REBASE \cite{DBLP:journals/corr/abs-2408-00724} effectively manages the exploitation-pruning trade-off in tree search, achieving better empirical results than sampling-based techniques and MCTS \cite{DBLP:journals/corr/abs-2408-00724}.
Reward models \cite{DBLP:conf/iclr/LightmanKBEBLLS24,DBLP:conf/acl/WangLSXDLCWS24} serve as crucial components in these approaches, existing in two primary forms: outcome-based and process-based. Outcome reward models \cite{DBLP:journals/corr/abs-2405-14333,DBLP:journals/corr/abs-2408-11791}, which evaluate final solutions through scoring, are especially valuable for Best-of-N selection strategies. Conversely, process reward models \cite{DBLP:journals/corr/abs-2405-14333,DBLP:conf/acl/WangLSXDLCWS24,DBLP:journals/corr/abs-2408-00724} analyze reasoning steps, making them particularly effective for guiding tree-based search.

\begin{figure*}[!t]
\vspace{-1em}
\begin{small}
\begin{tcolorbox}[colback=blue!10,colframe=blue!40!black,title=\textbf{\texttt{Takeways (Domain-Specific Applications)}}]
The standard approach for developing domain-specific SRMs involves adapting general models to specialized datasets. Consequently, many researchers construct custom datasets \cite{DBLP:journals/corr/abs-2309-06126,DBLP:conf/www/YangZKXHA24,DBLP:journals/corr/abs-2401-07950,DBLP:journals/corr/abs-2402-06852}, typically annotated using advanced LLMs (e.g., GPT-4) and are subsequently employed for continual pre-training or fine-tuning of models like LLaMA-2-7B \cite{DBLP:journals/corr/abs-2404-16621}. To ensure high-quality outputs, specialized annotation frameworks, such as the one used by SciGLM \cite{DBLP:journals/corr/abs-2401-07950}, have been developed.
For domains with substantial textual corpora, an effective strategy involves training a base model from scratch, followed by SFT~\cite{DBLP:journals/corr/abs-2310-15777}.
An alternative method for developing domain-specific SRMs involves a dual process of distilling general capabilities from LLMs, and systematically integrating specialized knowledge from domain corpora \cite{DBLP:conf/acl/YaoHWDW21}.
\end{tcolorbox}
\end{small}
\vspace{-1em}
\end{figure*}

\section{Domain-Specific Applications}
\label{sect:app}

While LRMs require broad knowledge, domain-specific SRMs focus more on expertise. Here, we emphasize popular applications in various domains.

\subsection{Healthcare}

Hippocrates~\cite{DBLP:journals/corr/abs-2404-16621} provides unrestricted access to datasets, codebase, models, and protocols\footnote{\url{https://cyberiada.github.io/Hippocrates/}}. It is trained on a medical corpus comprising Medical Guidelines, PMCPatients~\cite{DBLP:journals/corr/abs-2202-13876}, and PubMedQA-contexts~\cite{DBLP:conf/emnlp/JinDLCL19}, totaling 300M tokens. The Hippo series undergoes continuous pre-training, SFT, and RLHF. The fine-tuned versions of Mistral and Llama-2 compete with several 70B models in evaluation. For instance, Hippo-Mistral-7B achieves 59.9\% on MedQA, surpassing Meditron-70B \cite{DBLP:journals/corr/abs-2311-16079} (58.5\%).
BioMedLM \cite{DBLP:journals/corr/abs-2403-18421} is a 2.7B model pre-trained on PubMed \cite{DBLP:journals/corr/abs-2101-00027}. 
AdaLM~\cite{DBLP:conf/acl/YaoHWDW21} improves domain SRMs by performing continued training on a medical-focused SLM. The experiments show that the adaptation-then-distillation approach is most effective. 
MentalLLaMA \cite{DBLP:conf/www/YangZKXHA24} pioneers two key contributions: (1) the first IMHI dataset  for mental health analysis, and (2) the first open-source model capable of explainable analysis of social media content. 
HuatuoGPT-o1 \cite{DBLP:journals/corr/abs-2412-18925} introduces verifiable medical problems along with a medical verifier to assess the accuracy of model outputs. This verifiability allows progress in medical reasoning via a two-step method: (1) leveraging the verifier to guide the search for intricate reasoning paths, and (2) employing RL with rewards based on the verifier to further improve complex reasoning.

\subsection{Science}
SciGLM \cite{DBLP:journals/corr/abs-2401-07950} is a university-level scientific model that addresses data limitations by employing a self-reflective instruction annotation framework. Using GPT-4 \cite{DBLP:journals/corr/abs-2303-08774}, it produces step-by-step reasoning for unlabeled scientific questions through a three-stage process with structured prompts: (1) CoT prompt (guides the model to provide step-by-step reasoning), (2) reflective prompt (helps identify and correct errors in the reasoning process) and (3) answer integration (combines corrected solutions for clear and accurate output).
Additionally, Llemma \cite{DBLP:conf/iclr/AzerbayevSPSMJD24}, a scientific model adapted from CodeLlama \cite{DBLP:journals/corr/abs-2308-12950}, focuses on advanced mathematical reasoning. Through extended pre-training, its 7B model is further developed using 55B tokens from the newly constructed Proof-Pile-2 dataset. This dataset incorporates scientific publications, mathematical web content, and computational math resources. The enhanced model demonstrates superior performance on key mathematical benchmarks including MATH \cite{DBLP:conf/iclr/HendrycksBBZMSS21}, GSM8k \cite{DBLP:journals/corr/abs-2110-14168}, OCWCourses \cite{DBLP:conf/nips/LewkowyczADDMRS22}, MMLU-STEM \cite{DBLP:conf/iclr/HendrycksBBZMSS21}, and SAT, outperforming all similarly-sized models.
ChemLLM~\cite{DBLP:journals/corr/abs-2402-06852} is a chemistry SRM that employs its ChemData framework, which reformats chemical knowledge into conversational data. The model builds upon InternLM2-Base-7B~\cite{DBLP:journals/corr/abs-2403-17297}, first strengthening its foundational abilities through pre-training on 1.7M QA pairs from HuggingFace's multi-domain corpus. Subsequent SFT combines both ChemData and the multi-corpus to preserve broad capabilities while specializing in chemistry. ChemLLM demonstrates exceptional competence in interdisciplinary chemistry applications, matching GPT-4 \cite{DBLP:journals/corr/abs-2303-08774} in several domains while consistently surpassing GPT-3.5 \cite{DBLP:conf/nips/Ouyang0JAWMZASR22}. Notably, it achieves 92.6 on Mol2caption, nearly reaching GPT-4's performance level.
AstroLLaMA \cite{DBLP:journals/corr/abs-2309-06126} is a specialized model for astronomical applications. Built upon Llama-2-7B \cite{DBLP:journals/corr/abs-2307-09288}, the model undergoes extended pre-training using a curated collection of more than 300K astronomy abstracts sourced from arXiv\footnote{\url{https://www.kaggle.com/Cornell-University/arxiv}}. It supports various astronomy-related applications, including automated research paper summarization and the creation of conversational AI systems for astronomical research.

\subsection{Other Domains}
MindLLM \cite{DBLP:journals/corr/abs-2310-15777} is a bilingual (Chinese-English) model with dual-language pretraining: English capabilities are developed using the Pile dataset \cite{DBLP:journals/corr/abs-2101-00027}, while Chinese proficiency is cultivated through WuDao \cite{DBLP:journals/aiopen/YuanZDDLCZYT21}, CBook, and various Chinese web sources. This bilingual approach simultaneously enhances the model's capacity while mitigating catastrophic forgetting during specialization. The model exhibits notable strength in specialized domains through supervised fine-tuning: (1) legal applications: leveraging open legal datasets, scenario-based QA from LaW-GPT \cite{LAWGPT-zh}, and NLP legal tasks from DISC-LawLLM \cite{DBLP:journals/corr/abs-2309-11325}.
LaWGPT \cite{LAWGPT-zh} comprises a family of models designed to broaden legal vocabulary coverage.
These models undergo pre-training on extensive Chinese legal text corpora to improve foundational semantic comprehension capabilities within the legal domain.
Developed as a Chinese legal SRM, Lawyer LLaMA \cite{DBLP:journals/corr/abs-2305-15062} undergoes training on comprehensive legal datasets, enabling it to provide legal guidance, conduct case evaluations, and produce legal writings.
ChatLaw \cite{DBLP:journals/corr/abs-2306-16092} is a collection of open-source legal SRMs.
The series includes models like ChatLaw-13B and ChatLaw-33B, trained on extensive legal datasets comprising news, forum discussions, and judicial interpretations.
Additionally, the ChatLaw-Text2Vec variant employs a dataset of 930,000 court cases to develop a similarity-matching model.
(2) financial applications: utilizing EastMoney\footnote{\url{https://www.eastmoney.com/default.html}} as primary training data.
Fin-R1 \cite{liu2025} produces a high-quality CoT dataset, carefully distilled and filtered from multiple authoritative financial sources, tailored specifically for professional financial reasoning tasks.
Meanwhile, a financial SRM is specifically trained to meet the financial industry's key needs: decision-making support, numerical accuracy, etc.
As demonstrated by BloombergGPT \cite{DBLP:journals/corr/abs-2303-17564}, the training corpus combines general and financial content in equal proportions.
A key feature is the inclusion of a targeted 5-billion-token Bloomberg dataset, representing just 0.7\% of the overall training material.
This specialized data segment contributes significantly to improved results in financial benchmarking tests.

\section{Future Research Directions}

In this section, we highlight some possible future research directions on SRMs.

\noindent\textbf{Enhanced Distillation Techniques.}
While current distillation processes have successfully distilled LRMs into more manageable SRMs, there remains substantial room for improvement. Firstly, due to their small parameter sizes and capacities, the learning abilities of SRMs w.r.t. challenging tasks are often limited. Investigating methods that iteratively transfer complex reasoning knowledge from LRMs to SRMs is an interesting idea. Starting with basic reasoning tasks and progressively imparting complex CoT knowledge can lead to more robust SRMs. Secondly, while SFT-based distillation processes have achieved notable success, utilizing RL-in-the-loop mechanisms can help identify areas where SRMs may fall short, allowing for targeted adjustments to address these deficiencies. Finally, by integrating external knowledge sources (such as knowledge graphs) during the distillation process, researchers can enrich SRMs with additional contexts and background knowledge.

\noindent\textbf{Adaptive RL Strategies.}
To further enhance the capabilities of SRMs, adaptive RL strategies offer a promising research direction. For SRMs, it is worthwhile to study adaptive mechanisms to balance exploration and exploitation in RL, since SRMs have limited capacity for exploring alternatives too far away from reference models after SFT. By dynamically adjusting exploration rates based on task complexity and model performance, SRMs can better navigate the learning space for more effective reasoning. In addition, by optimizing rewards tailored to specific reasoning tasks, SRMs can learn more nuanced decision rules, leading to higher accuracy and efficiency. Finally, it is also interesting to develop continual learning frameworks that allow SRMs to incrementally update their knowledge and policies based on ongoing interactions and feedback. 

\noindent\textbf{Learning in Low-Resource Settings.}
Learning in low-resource settings is a critical frontier for SRMs. Many domains and applications suffer from a paucity of high-quality data, limited computational resources, and constrained environments. It is necessary to investigate strategies for leveraging CoT reasoning datasets from related domains to enrich low-resource datasets. Regarding computational resources, although there is abundant research on parameter-efficient learning for SFT~\cite{DBLP:conf/iclr/HuSWALWWC22,DBLP:conf/nips/DettmersPHZ23}, the question of whether and when parameter-efficient learning is useful for RL on SRMs still remains.

\noindent\textbf{Agent-based Efficient Inference.}
Although agent-based cooperation brings strong performance to SRMs in complex task reasoning, current multi-agent systems inevitably generate significant token overhead and incur higher operational costs, hindering their scalability for widespread adoption. It is necessary to integrate effortlessly into mainstream multi-agent systems while filtering out redundant or malicious communication \cite{DBLP:journals/corr/abs-2410-02506,agentdropout}.
Specifically, by optimizing the adjacency matrices of communication graphs, these systems can detect and remove redundant agents and cross-round interactions, thereby improving both token efficiency and task performance.

\section{Concluding Remarks}
In conclusion, this survey has provided a comprehensive overview of SRMs. The advancements in SRMs open up new avenues for deploying high-performance models in resource-constrained environments, which is crucial for both academic research and commercial applications. As SRMs continue to evolve, it is imperative for future research to not only enhance their reasoning capabilities but also to explore innovative ways to integrate SRMs into a broader range of NLP tasks. 

\newpage

\section*{Limitations}

While our survey provides an overview of SRMs and their applications, it is important to acknowledge several limitations inherent in this work. Firstly, the field of large and small reasoning models is rapidly evolving, with new models and techniques being developed continuously. Consequently, there is an inevitable delay in capturing the latest advancements, and certain cutting-edge developments may not be fully represented.
Secondly, our analysis primarily focuses on published literature, and there may be unpublished or proprietary advancements in SRMs that are not covered. This reliance on available publications may introduce a degree of publication bias, potentially limiting the scope of findings.
Finally, while we cover a broad range of domains in which SRMs have been applied, some niche or emerging domains may not be extensively discussed, reflecting the current state of research emphasis.

\section*{Ethical Considerations and Border Impact}

This work primarily reviews recent advances in SRMs and does not present direct ethical issues. However, it is important to acknowledge that many SRMs are distilled from larger models, raising concerns about potential biases and privacy issues being perpetuated. Addressing these concerns is beyond the scope of this paper. We recommend readers refer to related surveys~\cite{DBLP:journals/cee/KibriyaKSK24,DBLP:journals/cit/GaoLLY24,DBLP:journals/corr/abs-2307-16680} for more detailed discussions on these topics.

The exploration and development of SRMs have significant implications for the field of NLP and beyond. This survey has underscored the potential for SRMs to democratize access to advanced reasoning capabilities by reducing the computational resources required compared to larger models. This aspect is particularly relevant for institutions with limited hardware resources, fostering inclusivity in AI research and applications. Moreover, SRMs offer opportunities for sustainable AI research and deployment, as their reduced computational demands align with efforts to minimize the carbon footprint associated with large-scale AI models.


\appendix

\begin{table*}
\centering
\begin{small}
\begin{tabular}{l | ccc}
\hline
\textbf{Datset Key} & \textbf{Size} & \textbf{Annotator} & \textbf{Related Paper}\\
\hline
FreedomIntelligence/medical-o1-reasoning-SFT & 25.4K & GPT-4o & \cite{DBLP:journals/corr/abs-2412-18925}\\
FreedomIntelligence/Medical-R1-Distill-Data & 22K & DeepSeek-R1 & \cite{DBLP:journals/corr/abs-2412-18925}\\
FreedomIntelligence/Medical-R1-Distill-Data-Chinese & 17K & DeepSeek-R1 & \cite{DBLP:journals/corr/abs-2412-18925}\\
facebook/natural$\underline{\ }$reasoning & 2.8M & Llama3.3-70B-Instruct & \cite{yuan2025naturalreasoningreasoningwild28m}\\
Congliu/Chinese-DeepSeek-R1-Distill-data-110k & 110K & DeepSeek-R1 & -\\
open-r1/codeforces-cots & 47.8K & DeepSeek-R1 & -\\
open-r1/OpenR1-Math-220k & 220K & DeepSeek-R1 & -\\
SmallDoge/SmallThoughts & 50K & DeepSeek-R1 & -\\
GeneralReasoning/GeneralThought-323K & 323K & Multiple & -\\
GeneralReasoning/GeneralThought-195K & 195K & Multiple & -\\
open-thoughts/OpenThoughts-114k & 114K & DeepSeek-R1 & -\\
bespokelabs/Bespoke-Stratos-17k & 17K & DeepSeek-R1 & -\\
simplescaling/s1K-1.1 & 1K & DeepSeek-R1 & \cite{DBLP:journals/corr/abs-2501-19393}\\
EricLu/SCP-116K & 116K & o1-mini, QwQ-32B-Preview & \cite{DBLP:journals/corr/abs-2501-15587}\\
\hline
\end{tabular}
\end{small}
\caption{The list of open-source datsets with long CoT and output annotations by LRMs.}
\label{tab:data}
\end{table*}

\begin{table*}
\centering
\begin{small}
\begin{tabular}{l | cc}
\hline
\textbf{Repository} & \textbf{URL}\\
\hline
open-r1 & \url{https://github.com/huggingface/open-r1}\\
trl & \url{https://github.com/huggingface/trl}\\
verl~\cite{DBLP:journals/corr/abs-2409-19256} & \url{https://github.com/volcengine/verl}\\
ReaLHF~\cite{DBLP:journals/corr/abs-2406-14088} & \url{https://github.com/openpsi-project/ReaLHF}\\
AReaL~\cite{mei2025real} & \url{https://github.com/inclusionAI/AReaL}\\
deepscaler & \url{https://github.com/agentica-project/deepscaler}\\
Light-R1~\cite{wen2025light} & \url{https://github.com/Qihoo360/Light-R1}\\
Open-Reasoner-Zero~\cite{wen2025light} & \url{https://github.com/Open-Reasoner-Zero/Open-Reasoner-Zero}\\
OpenRLHF~\cite{DBLP:journals/corr/abs-2405-11143} & \url{https://github.com/OpenRLHF/OpenRLHF}\\
LLaMA-Factory~\cite{DBLP:journals/corr/abs-2403-13372} & \url{https://github.com/hiyouga/LLaMA-Factory}\\
\hline
\end{tabular}
\end{small}
\caption{The list of open-source repositories to support RL training for SRMs.}
\label{tab:repo}
\end{table*}


\begin{thebibliography}{166}
  \expandafter\ifx\csname natexlab\endcsname\relax\def\natexlab#1{#1}\fi
  
  \bibitem[{Acikgoz et~al.(2024)Acikgoz, Ince, Bench, Boz, Kesen, Erdem, and Erdem}]{DBLP:journals/corr/abs-2404-16621}
  Emre~Can Acikgoz, Osman~Batur Ince, Rayene Bench, Arda~Anil Boz, Ilker Kesen, Aykut Erdem, and Erkut Erdem. 2024.
  \newblock \href {https://doi.org/10.48550/arXiv.2404.16621} {Hippocrates: An open-source framework for advancing large language models in healthcare}.
  \newblock \emph{CoRR}, abs/2404.16621.
  
  \bibitem[{Ahn et~al.(2024)Ahn, Verma, Lou, Liu, Zhang, and Yin}]{DBLP:conf/eacl/AhnVLLZY24}
  Janice Ahn, Rishu Verma, Renze Lou, Di~Liu, Rui Zhang, and Wenpeng Yin. 2024.
  \newblock \href {https://aclanthology.org/2024.eacl-srw.17} {Large language models for mathematical reasoning: Progresses and challenges}.
  \newblock In \emph{Proceedings of the 18th Conference of the European Chapter of the Association for Computational Linguistics}, pages 225--237. Association for Computational Linguistics.
  
  \bibitem[{Amini et~al.(2024)Amini, Vieira, and Cotterell}]{DBLP:conf/acl/AminiVC24}
  Afra Amini, Tim Vieira, and Ryan Cotterell. 2024.
  \newblock \href {https://doi.org/10.18653/V1/2024.FINDINGS-ACL.592} {Direct preference optimization with an offset}.
  \newblock In \emph{Findings of the Association for Computational Linguistics, {ACL} 2024}, pages 9954--9972. Association for Computational Linguistics.
  
  \bibitem[{Ankner et~al.(2024)Ankner, Paul, Cui, Chang, and Ammanabrolu}]{DBLP:journals/corr/abs-2408-11791}
  Zachary Ankner, Mansheej Paul, Brandon Cui, Jonathan~D. Chang, and Prithviraj Ammanabrolu. 2024.
  \newblock \href {https://doi.org/10.48550/arXiv.2408.11791} {Critique-out-loud reward models}.
  \newblock \emph{CoRR}, abs/2408.11791.
  
  \bibitem[{Azerbayev et~al.(2024)Azerbayev, Schoelkopf, Paster, Santos, McAleer, Jiang, Deng, Biderman, and Welleck}]{DBLP:conf/iclr/AzerbayevSPSMJD24}
  Zhangir Azerbayev, Hailey Schoelkopf, Keiran Paster, Marco~Dos Santos, Stephen~Marcus McAleer, Albert~Q. Jiang, Jia Deng, Stella Biderman, and Sean Welleck. 2024.
  \newblock \href {https://openreview.net/forum?id=4WnqRR915j} {Llemma: An open language model for mathematics}.
  \newblock In \emph{The Twelfth International Conference on Learning Representations}.
  
  \bibitem[{Bai et~al.(2022)Bai, Kadavath, Kundu, Askell, Kernion, Jones, Chen, Goldie, Mirhoseini, McKinnon, Chen, Olsson, Olah, Hernandez, Drain, Ganguli, Li, Tran{-}Johnson, Perez, Kerr, Mueller, Ladish, Landau, Ndousse, Lukosiute, Lovitt, Sellitto, Elhage, Schiefer, Mercado, DasSarma, Lasenby, Larson, Ringer, Johnston, Kravec, Showk, Fort, Lanham, Telleen{-}Lawton, Conerly, Henighan, Hume, Bowman, Hatfield{-}Dodds, Mann, Amodei, Joseph, McCandlish, Brown, and Kaplan}]{DBLP:journals/corr/abs-2212-08073}
  Yuntao Bai, Saurav Kadavath, Sandipan Kundu, Amanda Askell, Jackson Kernion, Andy Jones, Anna Chen, Anna Goldie, Azalia Mirhoseini, Cameron McKinnon, Carol Chen, Catherine Olsson, Christopher Olah, Danny Hernandez, Dawn Drain, Deep Ganguli, Dustin Li, Eli Tran{-}Johnson, Ethan Perez, Jamie Kerr, Jared Mueller, Jeffrey Ladish, Joshua Landau, Kamal Ndousse, Kamile Lukosiute, Liane Lovitt, Michael Sellitto, Nelson Elhage, Nicholas Schiefer, Noem{\'{\i}} Mercado, Nova DasSarma, Robert Lasenby, Robin Larson, Sam Ringer, Scott Johnston, Shauna Kravec, Sheer~El Showk, Stanislav Fort, Tamera Lanham, Timothy Telleen{-}Lawton, Tom Conerly, Tom Henighan, Tristan Hume, Samuel~R. Bowman, Zac Hatfield{-}Dodds, Ben Mann, Dario Amodei, Nicholas Joseph, Sam McCandlish, Tom Brown, and Jared Kaplan. 2022.
  \newblock \href {https://doi.org/10.48550/ARXIV.2212.08073} {Constitutional {AI:} harmlessness from {AI} feedback}.
  \newblock \emph{CoRR}, abs/2212.08073.
  
  \bibitem[{Bi et~al.(2024)Bi, Han, Liu, Tang, and Wang}]{DBLP:journals/corr/abs-2412-09078}
  Zhenni Bi, Kai Han, Chuanjian Liu, Yehui Tang, and Yunhe Wang. 2024.
  \newblock \href {https://doi.org/10.48550/arXiv.2412.09078} {Forest-of-thought: Scaling test-time compute for enhancing {LLM} reasoning}.
  \newblock \emph{CoRR}, abs/2412.09078.
  
  \bibitem[{Bolton et~al.(2024)Bolton, Venigalla, Yasunaga, Hall, Xiong, Lee, Daneshjou, Frankle, Liang, Carbin, and Manning}]{DBLP:journals/corr/abs-2403-18421}
  Elliot Bolton, Abhinav Venigalla, Michihiro Yasunaga, David Hall, Betty Xiong, Tony Lee, Roxana Daneshjou, Jonathan Frankle, Percy Liang, Michael Carbin, and Christopher~D. Manning. 2024.
  \newblock \href {https://doi.org/10.48550/arXiv.2403.18421} {Biomedlm: {A} 2.7b parameter language model trained on biomedical text}.
  \newblock \emph{CoRR}, abs/2403.18421.
  
  \bibitem[{Brown et~al.(2020)Brown, Mann, Ryder, Subbiah, Kaplan, Dhariwal, Neelakantan, Shyam, Sastry, Askell, Agarwal, Herbert{-}Voss, Krueger, Henighan, Child, Ramesh, Ziegler, Wu, Winter, Hesse, Chen, Sigler, Litwin, Gray, Chess, Clark, Berner, McCandlish, Radford, Sutskever, and Amodei}]{DBLP:conf/nips/BrownMRSKDNSSAA20}
  Tom~B. Brown, Benjamin Mann, Nick Ryder, Melanie Subbiah, Jared Kaplan, Prafulla Dhariwal, Arvind Neelakantan, Pranav Shyam, Girish Sastry, Amanda Askell, Sandhini Agarwal, Ariel Herbert{-}Voss, Gretchen Krueger, Tom Henighan, Rewon Child, Aditya Ramesh, Daniel~M. Ziegler, Jeffrey Wu, Clemens Winter, Christopher Hesse, Mark Chen, Eric Sigler, Mateusz Litwin, Scott Gray, Benjamin Chess, Jack Clark, Christopher Berner, Sam McCandlish, Alec Radford, Ilya Sutskever, and Dario Amodei. 2020.
  \newblock \href {https://proceedings.neurips.cc/paper/2020/hash/1457c0d6bfcb4967418bfb8ac142f64a-Abstract.html} {Language models are few-shot learners}.
  \newblock In \emph{Advances in Neural Information Processing Systems 33: Annual Conference on Neural Information Processing Systems 2020}.
  
  \bibitem[{Cai et~al.(2024)Cai, Cao, Chen, Chen, Chen, Chen, Chen, Chen, Chen, Chu, Dong, Duan, Fan, Fei, Gao, Ge, Gu, Gu, Gui, Guo, Guo, He, Hu, Huang, Jiang, Jiao, Jin, Lei, Li, Li, Li, Li, Li, Li, Liu, Liu, Hong, Liu, Liu, Liu, Lv, Lv, Lv, Ma, Ma, Ma, Ning, Ouyang, Qiu, Qu, Shang, Shao, Song, Song, Sui, Sun, Sun, Tang, Wang, Wang, Wang, Wang, Wang, Wang, Wang, Wei, Weng, Wu, Xiong, Zhao, and et~al.}]{DBLP:journals/corr/abs-2403-17297}
  Zheng Cai, Maosong Cao, Haojiong Chen, Kai Chen, Keyu Chen, Xin Chen, Xun Chen, Zehui Chen, Zhi Chen, Pei Chu, Xiaoyi Dong, Haodong Duan, Qi~Fan, Zhaoye Fei, Yang Gao, Jiaye Ge, Chenya Gu, Yuzhe Gu, Tao Gui, Aijia Guo, Qipeng Guo, Conghui He, Yingfan Hu, Ting Huang, Tao Jiang, Penglong Jiao, Zhenjiang Jin, Zhikai Lei, Jiaxing Li, Jingwen Li, Linyang Li, Shuaibin Li, Wei Li, Yining Li, Hongwei Liu, Jiangning Liu, Jiawei Hong, Kaiwen Liu, Kuikun Liu, Xiaoran Liu, Chengqi Lv, Haijun Lv, Kai Lv, Li~Ma, Runyuan Ma, Zerun Ma, Wenchang Ning, Linke Ouyang, Jiantao Qiu, Yuan Qu, Fukai Shang, Yunfan Shao, Demin Song, Zifan Song, Zhihao Sui, Peng Sun, Yu~Sun, Huanze Tang, Bin Wang, Guoteng Wang, Jiaqi Wang, Jiayu Wang, Rui Wang, Yudong Wang, Ziyi Wang, Xingjian Wei, Qizhen Weng, Fan Wu, Yingtong Xiong, Xiaomeng Zhao, and et~al. 2024.
  \newblock \href {https://doi.org/10.48550/arXiv.2403.17297} {Internlm2 technical report}.
  \newblock \emph{CoRR}, abs/2403.17297.
  
  \bibitem[{Chan et~al.(2024)Chan, Chen, Su, Yu, Xue, Zhang, Fu, and Liu}]{DBLP:conf/iclr/ChanCSYXZF024}
  Chi{-}Min Chan, Weize Chen, Yusheng Su, Jianxuan Yu, Wei Xue, Shanghang Zhang, Jie Fu, and Zhiyuan Liu. 2024.
  \newblock \href {https://openreview.net/forum?id=FQepisCUWu} {Chateval: Towards better llm-based evaluators through multi-agent debate}.
  \newblock In \emph{The Twelfth International Conference on Learning Representations}.
  
  \bibitem[{Chen et~al.(2024{\natexlab{a}})Chen, Liao, Li, and Fan}]{DBLP:conf/emnlp/ChenL0024}
  Guoxin Chen, Minpeng Liao, Chengxi Li, and Kai Fan. 2024{\natexlab{a}}.
  \newblock \href {https://aclanthology.org/2024.findings-emnlp.463} {Step-level value preference optimization for mathematical reasoning}.
  \newblock In \emph{Findings of the Association for Computational Linguistics: {EMNLP} 2024}, pages 7889--7903. Association for Computational Linguistics.
  
  \bibitem[{Chen et~al.(2025)Chen, Ren, Chen, Yang, Sun, and Arik}]{DBLP:journals/corr/abs-2501-19306}
  Jiefeng Chen, Jie Ren, Xinyun Chen, Chengrun Yang, Ruoxi Sun, and Sercan~{\"{O}}. Arik. 2025.
  \newblock \href {https://doi.org/10.48550/arXiv.2501.19306} {{SETS:} leveraging self-verification and self-correction for improved test-time scaling}.
  \newblock \emph{CoRR}, abs/2501.19306.
  
  \bibitem[{Chen et~al.(2024{\natexlab{b}})Chen, Cai, Ji, Wang, Liu, Wang, Hou, and Wang}]{DBLP:journals/corr/abs-2412-18925}
  Junying Chen, Zhenyang Cai, Ke~Ji, Xidong Wang, Wanlong Liu, Rongsheng Wang, Jianye Hou, and Benyou Wang. 2024{\natexlab{b}}.
  \newblock \href {https://doi.org/10.48550/ARXIV.2412.18925} {Huatuogpt-o1, towards medical complex reasoning with llms}.
  \newblock \emph{CoRR}, abs/2412.18925.
  
  \bibitem[{Chen et~al.(2024{\natexlab{c}})Chen, Zhang, and Han}]{DBLP:conf/naacl/ChenZH24}
  Pei Chen, Shuai Zhang, and Boran Han. 2024{\natexlab{c}}.
  \newblock \href {https://doi.org/10.18653/v1/2024.findings-naacl.112} {Comm: Collaborative multi-agent, multi-reasoning-path prompting for complex problem solving}.
  \newblock In \emph{Findings of the Association for Computational Linguistics: {NAACL} 2024}, pages 1720--1738.
  
  \bibitem[{Chen et~al.(2024{\natexlab{d}})Chen, Lin, Sch{\"{a}}rli, and Zhou}]{DBLP:conf/iclr/ChenLSZ24}
  Xinyun Chen, Maxwell Lin, Nathanael Sch{\"{a}}rli, and Denny Zhou. 2024{\natexlab{d}}.
  \newblock \href {https://openreview.net/forum?id=KuPixIqPiq} {Teaching large language models to self-debug}.
  \newblock In \emph{The 12th International Conference on Learning Representations}.
  
  \bibitem[{Chen et~al.(2024{\natexlab{e}})Chen, Pan, Li, Ding, and Zhou}]{DBLP:journals/corr/abs-2411-19477}
  Yanxi Chen, Xuchen Pan, Yaliang Li, Bolin Ding, and Jingren Zhou. 2024{\natexlab{e}}.
  \newblock \href {https://doi.org/10.48550/arXiv.2411.19477} {A simple and provable scaling law for the test-time compute of large language models}.
  \newblock \emph{CoRR}, abs/2411.19477.
  
  \bibitem[{Chen et~al.(2023)Chen, Hern{\'{a}}ndez{-}Cano, Romanou, Bonnet, Matoba, Salvi, Pagliardini, Fan, K{\"{o}}pf, Mohtashami, Sallinen, Sakhaeirad, Swamy, Krawczuk, Bayazit, Marmet, Montariol, Hartley, Jaggi, and Bosselut}]{DBLP:journals/corr/abs-2311-16079}
  Zeming Chen, Alejandro Hern{\'{a}}ndez{-}Cano, Angelika Romanou, Antoine Bonnet, Kyle Matoba, Francesco Salvi, Matteo Pagliardini, Simin Fan, Andreas K{\"{o}}pf, Amirkeivan Mohtashami, Alexandre Sallinen, Alireza Sakhaeirad, Vinitra Swamy, Igor Krawczuk, Deniz Bayazit, Axel Marmet, Syrielle Montariol, Mary{-}Anne Hartley, Martin Jaggi, and Antoine Bosselut. 2023.
  \newblock \href {https://doi.org/10.48550/arXiv.2311.16079} {{MEDITRON-70B:} scaling medical pretraining for large language models}.
  \newblock \emph{CoRR}, abs/2311.16079.
  
  \bibitem[{Chen et~al.(2024{\natexlab{f}})Chen, White, Mooney, Payani, Su, and Sun}]{DBLP:conf/acl/ChenWMP0024}
  Ziru Chen, Michael White, Raymond~J. Mooney, Ali Payani, Yu~Su, and Huan Sun. 2024{\natexlab{f}}.
  \newblock \href {https://doi.org/10.18653/V1/2024.ACL-LONG.738} {When is tree search useful for {LLM} planning? it depends on the discriminator}.
  \newblock In \emph{Proceedings of the 62nd Annual Meeting of the Association for Computational Linguistics}, pages 13659--13678. Association for Computational Linguistics.
  
  \bibitem[{Chiang and Chen(2019)}]{DBLP:conf/naacl/ChiangC19}
  Ting{-}Rui Chiang and Yun{-}Nung Chen. 2019.
  \newblock \href {https://doi.org/10.18653/v1/n19-1272} {Semantically-aligned equation generation for solving and reasoning math word problems}.
  \newblock In \emph{Proceedings of the 2019 Conference of the North American Chapter of the Association for Computational Linguistics: Human Language Technologies}, pages 2656--2668.
  
  \bibitem[{Choi et~al.(2023)Choi, Fang, Wang, and Song}]{DBLP:conf/emnlp/ChoiF0S23}
  Sehyun Choi, Tianqing Fang, Zhaowei Wang, and Yangqiu Song. 2023.
  \newblock \href {https://doi.org/10.18653/v1/2023.emnlp-main.867} {{KCTS:} knowledge-constrained tree search decoding with token-level hallucination detection}.
  \newblock In \emph{Proceedings of the 2023 Conference on Empirical Methods in Natural Language Processing}, pages 14035--14053.
  
  \bibitem[{Cobbe et~al.(2021)Cobbe, Kosaraju, Bavarian, Chen, Jun, Kaiser, Plappert, Tworek, Hilton, Nakano, Hesse, and Schulman}]{DBLP:journals/corr/abs-2110-14168}
  Karl Cobbe, Vineet Kosaraju, Mohammad Bavarian, Mark Chen, Heewoo Jun, Lukasz Kaiser, Matthias Plappert, Jerry Tworek, Jacob Hilton, Reiichiro Nakano, Christopher Hesse, and John Schulman. 2021.
  \newblock \href {https://arxiv.org/abs/2110.14168} {Training verifiers to solve math word problems}.
  \newblock \emph{CoRR}, abs/2110.14168.
  
  \bibitem[{Cui et~al.(2023)Cui, Li, Yan, Chen, and Yuan}]{DBLP:journals/corr/abs-2306-16092}
  Jiaxi Cui, Zongjian Li, Yang Yan, Bohua Chen, and Li~Yuan. 2023.
  \newblock \href {https://doi.org/10.48550/arXiv.2306.16092} {Chatlaw: Open-source legal large language model with integrated external knowledge bases}.
  \newblock \emph{CoRR}, abs/2306.16092.
  
  \bibitem[{DeepSeek{-}AI(2025)}]{DBLP:journals/corr/abs-2501-12948}
  DeepSeek{-}AI. 2025.
  \newblock \href {https://doi.org/10.48550/ARXIV.2501.12948} {Deepseek-r1: Incentivizing reasoning capability in llms via reinforcement learning}.
  \newblock \emph{CoRR}, abs/2501.12948.
  
  \bibitem[{Dettmers et~al.(2023)Dettmers, Pagnoni, Holtzman, and Zettlemoyer}]{DBLP:conf/nips/DettmersPHZ23}
  Tim Dettmers, Artidoro Pagnoni, Ari Holtzman, and Luke Zettlemoyer. 2023.
  \newblock \href {http://papers.nips.cc/paper\_files/paper/2023/hash/1feb87871436031bdc0f2beaa62a049b-Abstract-Conference.html} {Qlora: Efficient finetuning of quantized llms}.
  \newblock In \emph{Advances in Neural Information Processing Systems 36: Annual Conference on Neural Information Processing Systems 2023}.
  
  \bibitem[{Du et~al.(2024)Du, Li, Torralba, Tenenbaum, and Mordatch}]{DBLP:conf/icml/Du00TM24}
  Yilun Du, Shuang Li, Antonio Torralba, Joshua~B. Tenenbaum, and Igor Mordatch. 2024.
  \newblock \href {https://openreview.net/forum?id=zj7YuTE4t8} {Improving factuality and reasoning in language models through multiagent debate}.
  \newblock In \emph{Forty-first International Conference on Machine Learning}.
  
  \bibitem[{Ethayarajh et~al.(2024)Ethayarajh, Xu, Muennighoff, Jurafsky, and Kiela}]{DBLP:journals/corr/abs-2402-01306}
  Kawin Ethayarajh, Winnie Xu, Niklas Muennighoff, Dan Jurafsky, and Douwe Kiela. 2024.
  \newblock \href {https://doi.org/10.48550/ARXIV.2402.01306} {{KTO:} model alignment as prospect theoretic optimization}.
  \newblock \emph{CoRR}, abs/2402.01306.
  
  \bibitem[{Fan et~al.(2023)Fan, Chen, Wang, and Huang}]{DBLP:journals/corr/abs-2307-16680}
  Mingyuan Fan, Cen Chen, Chengyu Wang, and Jun Huang. 2023.
  \newblock \href {https://doi.org/10.48550/ARXIV.2307.16680} {On the trustworthiness landscape of state-of-the-art generative models: {A} comprehensive survey}.
  \newblock \emph{CoRR}, abs/2307.16680.
  
  \bibitem[{Fu et~al.(2023{\natexlab{a}})Fu, Peng, Khot, and Lapata}]{DBLP:journals/corr/abs-2305-10142}
  Yao Fu, Hao Peng, Tushar Khot, and Mirella Lapata. 2023{\natexlab{a}}.
  \newblock \href {https://doi.org/10.48550/arXiv.2305.10142} {Improving language model negotiation with self-play and in-context learning from {AI} feedback}.
  \newblock \emph{CoRR}, abs/2305.10142.
  
  \bibitem[{Fu et~al.(2023{\natexlab{b}})Fu, Peng, Ou, Sabharwal, and Khot}]{DBLP:conf/icml/FuPOSK23}
  Yao Fu, Hao Peng, Litu Ou, Ashish Sabharwal, and Tushar Khot. 2023{\natexlab{b}}.
  \newblock \href {https://proceedings.mlr.press/v202/fu23d.html} {Specializing smaller language models towards multi-step reasoning}.
  \newblock In \emph{International Conference on Machine Learning, {ICML} 2023}, volume 202 of \emph{Proceedings of Machine Learning Research}, pages 10421--10430. {PMLR}.
  
  \bibitem[{Gandhi et~al.(2024)Gandhi, Lee, Grand, Liu, Cheng, Sharma, and Goodman}]{DBLP:journals/corr/abs-2404-03683}
  Kanishk Gandhi, Denise Lee, Gabriel Grand, Muxin Liu, Winson Cheng, Archit Sharma, and Noah~D. Goodman. 2024.
  \newblock \href {https://doi.org/10.48550/arXiv.2404.03683} {Stream of search (sos): Learning to search in language}.
  \newblock \emph{CoRR}, abs/2404.03683.
  
  \bibitem[{Gao et~al.(2021)Gao, Biderman, Black, Golding, Hoppe, Foster, Phang, He, Thite, Nabeshima, Presser, and Leahy}]{DBLP:journals/corr/abs-2101-00027}
  Leo Gao, Stella Biderman, Sid Black, Laurence Golding, Travis Hoppe, Charles Foster, Jason Phang, Horace He, Anish Thite, Noa Nabeshima, Shawn Presser, and Connor Leahy. 2021.
  \newblock \href {https://arxiv.org/abs/2101.00027} {The pile: An 800gb dataset of diverse text for language modeling}.
  \newblock \emph{CoRR}, abs/2101.00027.
  
  \bibitem[{Gao et~al.(2024)Gao, Liu, Lan, and Yang}]{DBLP:journals/cit/GaoLLY24}
  Zhengjie Gao, Xuanzi Liu, Yuanshuai Lan, and Zheng Yang. 2024.
  \newblock \href {http://cit.fer.hr/index.php/CIT/article/view/5778} {A brief survey on safety of large language models}.
  \newblock \emph{J. Comput. Inf. Technol.}, 32(1):47--64.
  
  \bibitem[{Giadikiaroglou et~al.(2024)Giadikiaroglou, Lymperaiou, Filandrianos, and Stamou}]{DBLP:conf/emnlp/GiadikiaroglouL24}
  Panagiotis Giadikiaroglou, Maria Lymperaiou, Giorgos Filandrianos, and Giorgos Stamou. 2024.
  \newblock \href {https://aclanthology.org/2024.emnlp-main.646} {Puzzle solving using reasoning of large language models: {A} survey}.
  \newblock In \emph{Proceedings of the 2024 Conference on Empirical Methods in Natural Language Processing}, pages 11574--11591. Association for Computational Linguistics.
  
  \bibitem[{Guan et~al.(2025)Guan, Zhang, Liu, Shang, Sun, Zhu, Yang, and Yang}]{DBLP:journals/corr/abs-2501-04519}
  Xinyu Guan, Li~Lyna Zhang, Yifei Liu, Ning Shang, Youran Sun, Yi~Zhu, Fan Yang, and Mao Yang. 2025.
  \newblock \href {https://doi.org/10.48550/ARXIV.2501.04519} {rstar-math: Small llms can master math reasoning with self-evolved deep thinking}.
  \newblock \emph{CoRR}, abs/2501.04519.
  
  \bibitem[{Guo et~al.(2024)Guo, Zhu, Yang, Xie, Dong, Zhang, Chen, Bi, Wu, Li, Luo, Xiong, and Liang}]{DBLP:journals/corr/abs-2401-14196}
  Daya Guo, Qihao Zhu, Dejian Yang, Zhenda Xie, Kai Dong, Wentao Zhang, Guanting Chen, Xiao Bi, Y.~Wu, Y.~K. Li, Fuli Luo, Yingfei Xiong, and Wenfeng Liang. 2024.
  \newblock \href {https://doi.org/10.48550/ARXIV.2401.14196} {Deepseek-coder: When the large language model meets programming - the rise of code intelligence}.
  \newblock \emph{CoRR}, abs/2401.14196.
  
  \bibitem[{Han et~al.(2024)Han, Gao, Liu, Zhang, and Zhang}]{DBLP:journals/corr/abs-2403-14608}
  Zeyu Han, Chao Gao, Jinyang Liu, Jeff Zhang, and Sai~Qian Zhang. 2024.
  \newblock \href {https://doi.org/10.48550/ARXIV.2403.14608} {Parameter-efficient fine-tuning for large models: {A} comprehensive survey}.
  \newblock \emph{CoRR}, abs/2403.14608.
  
  \bibitem[{Havrilla et~al.(2024)Havrilla, Raparthy, Nalmpantis, Dwivedi{-}Yu, Zhuravinskyi, Hambro, and Raileanu}]{DBLP:conf/icml/HavrillaRNDZHR24}
  Alexander Havrilla, Sharath~Chandra Raparthy, Christoforos Nalmpantis, Jane Dwivedi{-}Yu, Maksym Zhuravinskyi, Eric Hambro, and Roberta Raileanu. 2024.
  \newblock \href {https://openreview.net/forum?id=LH6R06NxdB} {Glore: When, where, and how to improve {LLM} reasoning via global and local refinements}.
  \newblock In \emph{Forty-first International Conference on Machine Learning}. OpenReview.net.
  
  \bibitem[{Hendrycks et~al.(2021)Hendrycks, Burns, Basart, Zou, Mazeika, Song, and Steinhardt}]{DBLP:conf/iclr/HendrycksBBZMSS21}
  Dan Hendrycks, Collin Burns, Steven Basart, Andy Zou, Mantas Mazeika, Dawn Song, and Jacob Steinhardt. 2021.
  \newblock \href {https://openreview.net/forum?id=d7KBjmI3GmQ} {Measuring massive multitask language understanding}.
  \newblock In \emph{9th International Conference on Learning Representations}.
  
  \bibitem[{Ho et~al.(2023)Ho, Schmid, and Yun}]{DBLP:conf/acl/HoSY23}
  Namgyu Ho, Laura Schmid, and Se{-}Young Yun. 2023.
  \newblock \href {https://doi.org/10.18653/V1/2023.ACL-LONG.830} {Large language models are reasoning teachers}.
  \newblock In \emph{Proceedings of the 61st Annual Meeting of the Association for Computational Linguistics}, pages 14852--14882. Association for Computational Linguistics.
  
  \bibitem[{Holt et~al.(2024)Holt, Luyten, and van~der Schaar}]{DBLP:conf/iclr/HoltLS24}
  Samuel Holt, Max~Ruiz Luyten, and Mihaela van~der Schaar. 2024.
  \newblock \href {https://openreview.net/forum?id=EhrzQwsV4K} {{L2MAC:} large language model automatic computer for extensive code generation}.
  \newblock In \emph{The Twelfth International Conference on Learning Representations}.
  
  \bibitem[{Hong et~al.(2024)Hong, Zhuge, Chen, Zheng, Cheng, Wang, Zhang, Wang, Yau, Lin, Zhou, Ran, Xiao, Wu, and Schmidhuber}]{DBLP:conf/iclr/HongZCZCWZWYLZR24}
  Sirui Hong, Mingchen Zhuge, Jonathan Chen, Xiawu Zheng, Yuheng Cheng, Jinlin Wang, Ceyao Zhang, Zili Wang, Steven Ka~Shing Yau, Zijuan Lin, Liyang Zhou, Chenyu Ran, Lingfeng Xiao, Chenglin Wu, and J{\"{u}}rgen Schmidhuber. 2024.
  \newblock \href {https://openreview.net/forum?id=VtmBAGCN7o} {Metagpt: Meta programming for {A} multi-agent collaborative framework}.
  \newblock In \emph{The Twelfth International Conference on Learning Representations}.
  
  \bibitem[{Hou et~al.(2025)Hou, Lv, Lu, Zhang, Li, Yao, Li, Tang, and Dong}]{DBLP:journals/corr/abs-2501-11651}
  Zhenyu Hou, Xin Lv, Rui Lu, Jiajie Zhang, Yujiang Li, Zijun Yao, Juanzi Li, Jie Tang, and Yuxiao Dong. 2025.
  \newblock \href {https://doi.org/10.48550/arXiv.2501.11651} {Advancing language model reasoning through reinforcement learning and inference scaling}.
  \newblock \emph{CoRR}, abs/2501.11651.
  
  \bibitem[{Hsieh et~al.(2023)Hsieh, Li, Yeh, Nakhost, Fujii, Ratner, Krishna, Lee, and Pfister}]{DBLP:conf/acl/HsiehLYNFRKLP23}
  Cheng{-}Yu Hsieh, Chun{-}Liang Li, Chih{-}Kuan Yeh, Hootan Nakhost, Yasuhisa Fujii, Alex Ratner, Ranjay Krishna, Chen{-}Yu Lee, and Tomas Pfister. 2023.
  \newblock \href {https://doi.org/10.18653/V1/2023.FINDINGS-ACL.507} {Distilling step-by-step! outperforming larger language models with less training data and smaller model sizes}.
  \newblock In \emph{Findings of the Association for Computational Linguistics: {ACL} 2023}, pages 8003--8017. Association for Computational Linguistics.
  
  \bibitem[{Hu et~al.(2022)Hu, Shen, Wallis, Allen{-}Zhu, Li, Wang, Wang, and Chen}]{DBLP:conf/iclr/HuSWALWWC22}
  Edward~J. Hu, Yelong Shen, Phillip Wallis, Zeyuan Allen{-}Zhu, Yuanzhi Li, Shean Wang, Lu~Wang, and Weizhu Chen. 2022.
  \newblock \href {https://openreview.net/forum?id=nZeVKeeFYf9} {Lora: Low-rank adaptation of large language models}.
  \newblock In \emph{The Tenth International Conference on Learning Representations}. OpenReview.net.
  
  \bibitem[{Hu et~al.(2024{\natexlab{a}})Hu, Wu, Wang, Xianyu, Zhang, and Cao}]{DBLP:journals/corr/abs-2405-11143}
  Jian Hu, Xibin Wu, Weixun Wang, Xianyu, Dehao Zhang, and Yu~Cao. 2024{\natexlab{a}}.
  \newblock \href {https://doi.org/10.48550/ARXIV.2405.11143} {Openrlhf: An easy-to-use, scalable and high-performance {RLHF} framework}.
  \newblock \emph{CoRR}, abs/2405.11143.
  
  \bibitem[{Hu et~al.(2024{\natexlab{b}})Hu, He, Wang, Zhao, Shao, and Nie}]{DBLP:conf/aaai/HuHWZSN24}
  Linmei Hu, Hongyu He, Duokang Wang, Ziwang Zhao, Yingxia Shao, and Liqiang Nie. 2024{\natexlab{b}}.
  \newblock \href {https://doi.org/10.1609/AAAI.V38I16.29782} {{LLM} vs small model? large language model based text augmentation enhanced personality detection model}.
  \newblock In \emph{Thirty-Eighth {AAAI} Conference on Artificial Intelligence}, pages 18234--18242. {AAAI} Press.
  
  \bibitem[{Hu et~al.(2024{\natexlab{c}})Hu, Shen, Zhang, and Tao}]{DBLP:conf/iclr/Hu00T24}
  Shengchao Hu, Li~Shen, Ya~Zhang, and Dacheng Tao. 2024{\natexlab{c}}.
  \newblock \href {https://openreview.net/forum?id=Qox9rO0kN0} {Learning multi-agent communication from graph modeling perspective}.
  \newblock In \emph{The Twelfth International Conference on Learning Representations}.
  
  \bibitem[{Huang and Chang(2023)}]{DBLP:conf/acl/0009C23}
  Jie Huang and Kevin~Chen{-}Chuan Chang. 2023.
  \newblock \href {https://doi.org/10.18653/V1/2023.FINDINGS-ACL.67} {Towards reasoning in large language models: {A} survey}.
  \newblock In \emph{Findings of the Association for Computational Linguistics: {ACL} 2023}, pages 1049--1065. Association for Computational Linguistics.
  
  \bibitem[{Huang et~al.(2023)Huang, Tao, An, Zhang, Jiang, Chen, Wu, and Feng}]{DBLP:journals/corr/abs-2305-15062}
  Quzhe Huang, Mingxu Tao, Zhenwei An, Chen Zhang, Cong Jiang, Zhibin Chen, Zirui Wu, and Yansong Feng. 2023.
  \newblock \href {https://doi.org/10.48550/arXiv.2305.15062} {Lawyer llama technical report}.
  \newblock \emph{CoRR}, abs/2305.15062.
  
  \bibitem[{Hui et~al.(2024)Hui, Yang, Cui, Yang, Liu, Zhang, Liu, Zhang, Yu, Dang, Yang, Men, Huang, Ren, Ren, Zhou, and Lin}]{DBLP:journals/corr/abs-2409-12186}
  Binyuan Hui, Jian Yang, Zeyu Cui, Jiaxi Yang, Dayiheng Liu, Lei Zhang, Tianyu Liu, Jiajun Zhang, Bowen Yu, Kai Dang, An~Yang, Rui Men, Fei Huang, Xingzhang Ren, Xuancheng Ren, Jingren Zhou, and Junyang Lin. 2024.
  \newblock \href {https://doi.org/10.48550/ARXIV.2409.12186} {Qwen2.5-coder technical report}.
  \newblock \emph{CoRR}, abs/2409.12186.
  
  \bibitem[{Hwang et~al.(2024)Hwang, Kim, Kim, Ye, and Seo}]{DBLP:journals/corr/abs-2404-10346}
  Hyeonbin Hwang, Doyoung Kim, Seungone Kim, Seonghyeon Ye, and Minjoon Seo. 2024.
  \newblock \href {https://doi.org/10.48550/ARXIV.2404.10346} {Self-explore to avoid the pit: Improving the reasoning capabilities of language models with fine-grained rewards}.
  \newblock \emph{CoRR}, abs/2404.10346.
  
  \bibitem[{Jiang et~al.(2023)Jiang, Ren, and Lin}]{DBLP:conf/acl/Jiang0L23}
  Dongfu Jiang, Xiang Ren, and Bill~Yuchen Lin. 2023.
  \newblock \href {https://doi.org/10.18653/v1/2023.acl-long.792} {Llm-blender: Ensembling large language models with pairwise ranking and generative fusion}.
  \newblock In \emph{Proceedings of the 61st Annual Meeting of the Association for Computational Linguistics}, pages 14165--14178.
  
  \bibitem[{Jiang et~al.(2024)Jiang, Wang, Shen, Kim, and Kim}]{DBLP:journals/corr/abs-2406-00515}
  Juyong Jiang, Fan Wang, Jiasi Shen, Sungju Kim, and Sunghun Kim. 2024.
  \newblock \href {https://doi.org/10.48550/ARXIV.2406.00515} {A survey on large language models for code generation}.
  \newblock \emph{CoRR}, abs/2406.00515.
  
  \bibitem[{Jin et~al.(2019)Jin, Dhingra, Liu, Cohen, and Lu}]{DBLP:conf/emnlp/JinDLCL19}
  Qiao Jin, Bhuwan Dhingra, Zhengping Liu, William~W. Cohen, and Xinghua Lu. 2019.
  \newblock \href {https://doi.org/10.18653/v1/D19-1259} {Pubmedqa: {A} dataset for biomedical research question answering}.
  \newblock In \emph{Proceedings of the 2019 Conference on Empirical Methods in Natural Language Processing and the 9th International Joint Conference on Natural Language Processing}, pages 2567--2577.
  
  \bibitem[{Kazemnejad et~al.(2024)Kazemnejad, Aghajohari, Portelance, Sordoni, Reddy, Courville, and Roux}]{DBLP:journals/corr/abs-2410-01679}
  Amirhossein Kazemnejad, Milad Aghajohari, Eva Portelance, Alessandro Sordoni, Siva Reddy, Aaron~C. Courville, and Nicolas~Le Roux. 2024.
  \newblock \href {https://doi.org/10.48550/ARXIV.2410.01679} {Vineppo: Unlocking {RL} potential for {LLM} reasoning through refined credit assignment}.
  \newblock \emph{CoRR}, abs/2410.01679.
  
  \bibitem[{Khattab et~al.(2023)Khattab, Singhvi, Maheshwari, Zhang, Santhanam, Vardhamanan, Haq, Sharma, Joshi, Moazam, Miller, Zaharia, and Potts}]{DBLP:journals/corr/abs-2310-03714}
  Omar Khattab, Arnav Singhvi, Paridhi Maheshwari, Zhiyuan Zhang, Keshav Santhanam, Sri Vardhamanan, Saiful Haq, Ashutosh Sharma, Thomas~T. Joshi, Hanna Moazam, Heather Miller, Matei Zaharia, and Christopher Potts. 2023.
  \newblock \href {https://doi.org/10.48550/arXiv.2310.03714} {Dspy: Compiling declarative language model calls into self-improving pipelines}.
  \newblock \emph{CoRR}, abs/2310.03714.
  
  \bibitem[{Kibriya et~al.(2024)Kibriya, Khan, Siddiqa, and Khan}]{DBLP:journals/cee/KibriyaKSK24}
  Hareem Kibriya, Wazir~Zada Khan, Ayesha Siddiqa, and Muhammad~Khurrum Khan. 2024.
  \newblock \href {https://doi.org/10.1016/J.COMPELECENG.2024.109698} {Privacy issues in large language models: {A} survey}.
  \newblock \emph{Comput. Electr. Eng.}, 120:109698.
  
  \bibitem[{Kim et~al.(2024)Kim, Mitra, Chen, Rahman, and Zhang}]{DBLP:conf/eacl/KimMCRZ24}
  Hannah Kim, Kushan Mitra, Rafael~Li Chen, Sajjadur Rahman, and Dan Zhang. 2024.
  \newblock \href {https://aclanthology.org/2024.eacl-demo.18} {Meganno+: {A} human-llm collaborative annotation system}.
  \newblock In \emph{Proceedings of the 18th Conference of the European Chapter of the Association for Computational Linguistics}, pages 168--176. Association for Computational Linguistics.
  
  \bibitem[{Kumar et~al.(2024)Kumar, Zhuang, Agarwal, Su, Co{-}Reyes, Singh, Baumli, Iqbal, Bishop, Roelofs, Zhang, McKinney, Shrivastava, Paduraru, Tucker, Precup, Behbahani, and Faust}]{DBLP:journals/corr/abs-2409-12917}
  Aviral Kumar, Vincent Zhuang, Rishabh Agarwal, Yi~Su, John~D. Co{-}Reyes, Avi Singh, Kate Baumli, Shariq Iqbal, Colton Bishop, Rebecca Roelofs, Lei~M. Zhang, Kay McKinney, Disha Shrivastava, Cosmin Paduraru, George Tucker, Doina Precup, Feryal M.~P. Behbahani, and Aleksandra Faust. 2024.
  \newblock \href {https://doi.org/10.48550/ARXIV.2409.12917} {Training language models to self-correct via reinforcement learning}.
  \newblock \emph{CoRR}, abs/2409.12917.
  
  \bibitem[{Kwon et~al.(2024)Kwon, Palo, and Johns}]{DBLP:journals/ral/KwonPJ24}
  Teyun Kwon, Norman~Di Palo, and Edward Johns. 2024.
  \newblock \href {https://doi.org/10.1109/LRA.2024.3410155} {Language models as zero-shot trajectory generators}.
  \newblock \emph{{IEEE} Robotics Autom. Lett.}, 9(7):6728--6735.
  
  \bibitem[{Lample et~al.(2022)Lample, Lacroix, Lachaux, Rodriguez, Hayat, Lavril, Ebner, and Martinet}]{DBLP:conf/nips/LampleLLRHLEM22}
  Guillaume Lample, Timoth{\'{e}}e Lacroix, Marie{-}Anne Lachaux, Aur{\'{e}}lien Rodriguez, Amaury Hayat, Thibaut Lavril, Gabriel Ebner, and Xavier Martinet. 2022.
  \newblock \href {http://papers.nips.cc/paper\_files/paper/2022/hash/a8901c5e85fb8e1823bbf0f755053672-Abstract-Conference.html} {Hypertree proof search for neural theorem proving}.
  \newblock In \emph{Advances in Neural Information Processing Systems 35: Annual Conference on Neural Information Processing Systems 2022}.
  
  \bibitem[{Lee et~al.(2025)Lee, Fischer, Wu, Marwood, Baluja, Schuurmans, and Chen}]{DBLP:journals/corr/abs-2501-09891}
  Kuang{-}Huei Lee, Ian Fischer, Yueh{-}Hua Wu, Dave Marwood, Shumeet Baluja, Dale Schuurmans, and Xinyun Chen. 2025.
  \newblock \href {https://doi.org/10.48550/arXiv.2501.09891} {Evolving deeper {LLM} thinking}.
  \newblock \emph{CoRR}, abs/2501.09891.
  
  \bibitem[{Lewkowycz et~al.(2022)Lewkowycz, Andreassen, Dohan, Dyer, Michalewski, Ramasesh, Slone, Anil, Schlag, Gutman{-}Solo, Wu, Neyshabur, Gur{-}Ari, and Misra}]{DBLP:conf/nips/LewkowyczADDMRS22}
  Aitor Lewkowycz, Anders Andreassen, David Dohan, Ethan Dyer, Henryk Michalewski, Vinay~V. Ramasesh, Ambrose Slone, Cem Anil, Imanol Schlag, Theo Gutman{-}Solo, Yuhuai Wu, Behnam Neyshabur, Guy Gur{-}Ari, and Vedant Misra. 2022.
  \newblock \href {http://papers.nips.cc/paper\_files/paper/2022/hash/18abbeef8cfe9203fdf9053c9c4fe191-Abstract-Conference.html} {Solving quantitative reasoning problems with language models}.
  \newblock In \emph{Advances in Neural Information Processing Systems 35: Annual Conference on Neural Information Processing Systems 2022}.
  
  \bibitem[{Li et~al.(2024)Li, Dong, Xue, Peng, Wang, and Liu}]{DBLP:journals/corr/abs-2407-04078}
  Chengpeng Li, Guanting Dong, Mingfeng Xue, Ru~Peng, Xiang Wang, and Dayiheng Liu. 2024.
  \newblock \href {https://doi.org/10.48550/ARXIV.2407.04078} {Dotamath: Decomposition of thought with code assistance and self-correction for mathematical reasoning}.
  \newblock \emph{CoRR}, abs/2407.04078.
  
  \bibitem[{Li(2024)}]{DBLP:conf/emnlp/Li24}
  Jiyi Li. 2024.
  \newblock \href {https://aclanthology.org/2024.emnlp-main.874} {Human-llm hybrid text answer aggregation for crowd annotations}.
  \newblock In \emph{Proceedings of the 2024 Conference on Empirical Methods in Natural Language Processing}, pages 15609--15622. Association for Computational Linguistics.
  
  \bibitem[{Li et~al.(2023)Li, Hessel, Yu, Ren, Chang, and Choi}]{DBLP:conf/acl/LiHYRC023}
  Liunian~Harold Li, Jack Hessel, Youngjae Yu, Xiang Ren, Kai{-}Wei Chang, and Yejin Choi. 2023.
  \newblock \href {https://doi.org/10.18653/V1/2023.ACL-LONG.150} {Symbolic chain-of-thought distillation: Small models can also "think" step-by-step}.
  \newblock In \emph{Proceedings of the 61st Annual Meeting of the Association for Computational Linguistics}, pages 2665--2679. Association for Computational Linguistics.
  
  \bibitem[{Liang et~al.(2024)Liang, He, Jiao, Wang, Wang, Wang, Yang, Shi, and Tu}]{DBLP:conf/emnlp/Liang0JW00Y0T24}
  Tian Liang, Zhiwei He, Wenxiang Jiao, Xing Wang, Yan Wang, Rui Wang, Yujiu Yang, Shuming Shi, and Zhaopeng Tu. 2024.
  \newblock \href {https://aclanthology.org/2024.emnlp-main.992} {Encouraging divergent thinking in large language models through multi-agent debate}.
  \newblock In \emph{Proceedings of the 2024 Conference on Empirical Methods in Natural Language Processing}, pages 17889--17904.
  
  \bibitem[{Lightman et~al.(2024)Lightman, Kosaraju, Burda, Edwards, Baker, Lee, Leike, Schulman, Sutskever, and Cobbe}]{DBLP:conf/iclr/LightmanKBEBLLS24}
  Hunter Lightman, Vineet Kosaraju, Yuri Burda, Harrison Edwards, Bowen Baker, Teddy Lee, Jan Leike, John Schulman, Ilya Sutskever, and Karl Cobbe. 2024.
  \newblock \href {https://openreview.net/forum?id=v8L0pN6EOi} {Let's verify step by step}.
  \newblock In \emph{The Twelfth International Conference on Learning Representations}. OpenReview.net.
  
  \bibitem[{Liu et~al.(2022)Liu, Dou, Li, Xu, and Liu}]{DBLP:conf/cogsci/LiuD0XL22}
  Yuntao Liu, Yong Dou, Yuan Li, Xinhai Xu, and Donghong Liu. 2022.
  \newblock \href {https://escholarship.org/uc/item/11z508nc} {Temporal dynamic weighted graph convolution for multi-agent reinforcement learning}.
  \newblock In \emph{Proceedings of the 44th Annual Meeting of the Cognitive Science Society, CogSci 2022, Toronto, ON, Canada, July 27-30, 2022}. cognitivesciencesociety.org.
  
  \bibitem[{Liu et~al.(2025)Liu, Guo, Lou, Zeng, Niu, Wang, Xu, Cai, Yang, Zhao, Li, Xu, Chen, Chen, Bai, and Zhang}]{liu2025}
  Zhaowei Liu, Xin Guo, Fangqi Lou, Lingfeng Zeng, Jinyi Niu, Zixuan Wang, Jiajie Xu, Weige Cai, Ziwei Yang, Xueqian Zhao, Chao Li, Sheng Xu, Dezhi Chen, Yun Chen, Zuo Bai, and Liwen Zhang. 2025.
  \newblock \href {https://arxiv.org/abs/2503.16252} {Fin-r1: A large language model for financial reasoning through reinforcement learning}.
  \newblock \emph{CoRR}, abs/2503.16252.
  
  \bibitem[{Liu et~al.(2023)Liu, Zhang, Li, Liu, and Yang}]{DBLP:journals/corr/abs-2310-02170}
  Zijun Liu, Yanzhe Zhang, Peng Li, Yang Liu, and Diyi Yang. 2023.
  \newblock \href {https://doi.org/10.48550/arXiv.2310.02170} {Dynamic llm-agent network: An llm-agent collaboration framework with agent team optimization}.
  \newblock \emph{CoRR}, abs/2310.02170.
  
  \bibitem[{Long(2023)}]{DBLP:journals/corr/abs-2305-08291}
  Jieyi Long. 2023.
  \newblock \href {https://doi.org/10.48550/arXiv.2305.08291} {Large language model guided tree-of-thought}.
  \newblock \emph{CoRR}, abs/2305.08291.
  
  \bibitem[{Lozhkov et~al.(2024)Lozhkov, Li, Allal, Cassano, Lamy{-}Poirier, Tazi, Tang, Pykhtar, Liu, Wei, Liu, Tian, Kocetkov, Zucker, Belkada, Wang, Liu, Abulkhanov, Paul, Li, Li, Risdal, Li, Zhu, Zhuo, Zheltonozhskii, Dade, Yu, Krau{\ss}, Jain, Su, He, Dey, Abati, Chai, Muennighoff, Tang, Oblokulov, Akiki, Marone, Mou, Mishra, Gu, Hui, Dao, Zebaze, Dehaene, Patry, Xu, McAuley, Hu, Scholak, Paquet, Robinson, Anderson, Chapados, and et~al.}]{DBLP:journals/corr/abs-2402-19173}
  Anton Lozhkov, Raymond Li, Loubna~Ben Allal, Federico Cassano, Joel Lamy{-}Poirier, Nouamane Tazi, Ao~Tang, Dmytro Pykhtar, Jiawei Liu, Yuxiang Wei, Tianyang Liu, Max Tian, Denis Kocetkov, Arthur Zucker, Younes Belkada, Zijian Wang, Qian Liu, Dmitry Abulkhanov, Indraneil Paul, Zhuang Li, Wen{-}Ding Li, Megan Risdal, Jia Li, Jian Zhu, Terry~Yue Zhuo, Evgenii Zheltonozhskii, Nii Osae~Osae Dade, Wenhao Yu, Lucas Krau{\ss}, Naman Jain, Yixuan Su, Xuanli He, Manan Dey, Edoardo Abati, Yekun Chai, Niklas Muennighoff, Xiangru Tang, Muhtasham Oblokulov, Christopher Akiki, Marc Marone, Chenghao Mou, Mayank Mishra, Alex Gu, Binyuan Hui, Tri Dao, Armel Zebaze, Olivier Dehaene, Nicolas Patry, Canwen Xu, Julian~J. McAuley, Han Hu, Torsten Scholak, S{\'{e}}bastien Paquet, Jennifer Robinson, Carolyn~Jane Anderson, Nicolas Chapados, and et~al. 2024.
  \newblock \href {https://doi.org/10.48550/ARXIV.2402.19173} {Starcoder 2 and the stack v2: The next generation}.
  \newblock \emph{CoRR}, abs/2402.19173.
  
  \bibitem[{Lu et~al.(2025)Lu, Tan, Xu, Yao, Qu, Chu, Xu, and Qi}]{DBLP:journals/corr/abs-2501-15587}
  Dakuan Lu, Xiaoyu Tan, Rui Xu, Tianchu Yao, Chao Qu, Wei Chu, Yinghui Xu, and Yuan Qi. 2025.
  \newblock \href {https://doi.org/10.48550/ARXIV.2501.15587} {{SCP-116K:} {A} high-quality problem-solution dataset and a generalized pipeline for automated extraction in the higher education science domain}.
  \newblock \emph{CoRR}, abs/2501.15587.
  
  \bibitem[{Luo et~al.(2023)Luo, Sun, Xu, Zhao, Lou, Tao, Geng, Lin, Chen, and Zhang}]{DBLP:journals/corr/abs-2308-09583}
  Haipeng Luo, Qingfeng Sun, Can Xu, Pu~Zhao, Jianguang Lou, Chongyang Tao, Xiubo Geng, Qingwei Lin, Shifeng Chen, and Dongmei Zhang. 2023.
  \newblock \href {https://doi.org/10.48550/ARXIV.2308.09583} {Wizardmath: Empowering mathematical reasoning for large language models via reinforced evol-instruct}.
  \newblock \emph{CoRR}, abs/2308.09583.
  
  \bibitem[{Madaan et~al.(2023)Madaan, Tandon, Gupta, Hallinan, Gao, Wiegreffe, Alon, Dziri, Prabhumoye, Yang, Gupta, Majumder, Hermann, Welleck, Yazdanbakhsh, and Clark}]{DBLP:conf/nips/MadaanTGHGW0DPY23}
  Aman Madaan, Niket Tandon, Prakhar Gupta, Skyler Hallinan, Luyu Gao, Sarah Wiegreffe, Uri Alon, Nouha Dziri, Shrimai Prabhumoye, Yiming Yang, Shashank Gupta, Bodhisattwa~Prasad Majumder, Katherine Hermann, Sean Welleck, Amir Yazdanbakhsh, and Peter Clark. 2023.
  \newblock \href {http://papers.nips.cc/paper\_files/paper/2023/hash/91edff07232fb1b55a505a9e9f6c0ff3-Abstract-Conference.html} {Self-refine: Iterative refinement with self-feedback}.
  \newblock In \emph{Advances in Neural Information Processing Systems 36: Annual Conference on Neural Information Processing Systems 2023}.
  
  \bibitem[{Magister et~al.(2023)Magister, Mallinson, Ad{\'{a}}mek, Malmi, and Severyn}]{DBLP:conf/acl/MagisterMAMS23}
  Lucie~Charlotte Magister, Jonathan Mallinson, Jakub Ad{\'{a}}mek, Eric Malmi, and Aliaksei Severyn. 2023.
  \newblock \href {https://doi.org/10.18653/V1/2023.ACL-SHORT.151} {Teaching small language models to reason}.
  \newblock In \emph{Proceedings of the 61st Annual Meeting of the Association for Computational Linguistics}, pages 1773--1781. Association for Computational Linguistics.
  
  \bibitem[{Mei et~al.(2024)Mei, Fu, Li, Wang, Zhang, and Wu}]{DBLP:journals/corr/abs-2406-14088}
  Zhiyu Mei, Wei Fu, Kaiwei Li, Guangju Wang, Huanchen Zhang, and Yi~Wu. 2024.
  \newblock \href {https://doi.org/10.48550/ARXIV.2406.14088} {Realhf: Optimized {RLHF} training for large language models through parameter reallocation}.
  \newblock \emph{CoRR}, abs/2406.14088.
  
  \bibitem[{Mei et~al.(2025)Mei, Fu, Li, Wang, Zhang, and Wu}]{mei2025real}
  Zhiyu Mei, Wei Fu, Kaiwei Li, Guangju Wang, Huanchen Zhang, and Yi~Wu. 2025.
  \newblock Real: Efficient rlhf training of large language models with parameter reallocation.
  \newblock In \emph{Proceedings of the Eighth Conference on Machine Learning and Systems}. mlsys.org.
  
  \bibitem[{Meng et~al.(2024)Meng, Xia, and Chen}]{DBLP:conf/nips/0001X024}
  Yu~Meng, Mengzhou Xia, and Danqi Chen. 2024.
  \newblock \href {http://papers.nips.cc/paper\_files/paper/2024/hash/e099c1c9699814af0be873a175361713-Abstract-Conference.html} {Simpo: Simple preference optimization with a reference-free reward}.
  \newblock In \emph{Advances in Neural Information Processing Systems 38: Annual Conference on Neural Information Processing Systems 2024}.
  
  \bibitem[{Mikulov{\'{a}} et~al.(2022)Mikulov{\'{a}}, Straka, Step{\'{a}}nek, Step{\'{a}}nkov{\'{a}}, and Hajic}]{DBLP:conf/lrec/MikulovaSSSH22}
  Marie Mikulov{\'{a}}, Milan Straka, Jan Step{\'{a}}nek, Barbora Step{\'{a}}nkov{\'{a}}, and Jan Hajic. 2022.
  \newblock \href {https://aclanthology.org/2022.lrec-1.312} {Quality and efficiency of manual annotation: Pre-annotation bias}.
  \newblock In \emph{Proceedings of the Thirteenth Language Resources and Evaluation Conference}, pages 2909--2918. European Language Resources Association.
  
  \bibitem[{Motwani et~al.(2024)Motwani, Smith, Das, Rybchuk, Torr, Laptev, Pizzati, Clark, and de~Witt}]{DBLP:journals/corr/abs-2412-01928}
  Sumeet~Ramesh Motwani, Chandler Smith, Rocktim~Jyoti Das, Markian Rybchuk, Philip H.~S. Torr, Ivan Laptev, Fabio Pizzati, Ronald Clark, and Christian~Schr{\"{o}}der de~Witt. 2024.
  \newblock \href {https://doi.org/10.48550/ARXIV.2412.01928} {{MALT:} improving reasoning with multi-agent {LLM} training}.
  \newblock \emph{CoRR}, abs/2412.01928.
  
  \bibitem[{Movva et~al.(2024)Movva, Koh, and Pierson}]{DBLP:conf/emnlp/MovvaKP24}
  Rajiv Movva, Pang~Wei Koh, and Emma Pierson. 2024.
  \newblock \href {https://aclanthology.org/2024.emnlp-main.511} {Annotation alignment: Comparing {LLM} and human annotations of conversational safety}.
  \newblock In \emph{Proceedings of the 2024 Conference on Empirical Methods in Natural Language Processing}, pages 9048--9062. Association for Computational Linguistics.
  
  \bibitem[{Muennighoff et~al.(2025)Muennighoff, Yang, Shi, Li, Fei{-}Fei, Hajishirzi, Zettlemoyer, Liang, Cand{\`{e}}s, and Hashimoto}]{DBLP:journals/corr/abs-2501-19393}
  Niklas Muennighoff, Zitong Yang, Weijia Shi, Xiang~Lisa Li, Li~Fei{-}Fei, Hannaneh Hajishirzi, Luke Zettlemoyer, Percy Liang, Emmanuel~J. Cand{\`{e}}s, and Tatsunori Hashimoto. 2025.
  \newblock \href {https://doi.org/10.48550/ARXIV.2501.19393} {s1: Simple test-time scaling}.
  \newblock \emph{CoRR}, abs/2501.19393.
  
  \bibitem[{Nguyen et~al.(2023)Nguyen, Ting, Ciuca, O'Neill, Sun, Jablonska, Kruk, Perkowski, Miller, Li, Peek, Iyer, R{\'{o}}zanski, Khetarpal, Zaman, Brodrick, M{\'{e}}ndez, Bui, Goodman, Accomazzi, Naiman, Cranney, Schawinski, and UniverseTBD}]{DBLP:journals/corr/abs-2309-06126}
  Tuan~Dung Nguyen, Yuan{-}Sen Ting, Ioana Ciuca, Charlie O'Neill, Zechang Sun, Maja Jablonska, Sandor Kruk, Ernest Perkowski, Jack~W. Miller, Jason Li, Josh Peek, Kartheik Iyer, Tomasz R{\'{o}}zanski, Pranav Khetarpal, Sharaf Zaman, David Brodrick, Sergio J.~Rodr{\'{\i}}guez M{\'{e}}ndez, Thang~D. Bui, Alyssa Goodman, Alberto Accomazzi, Jill~P. Naiman, Jesse Cranney, Kevin Schawinski, and UniverseTBD. 2023.
  \newblock \href {https://doi.org/10.48550/arXiv.2309.06126} {Astrollama: Towards specialized foundation models in astronomy}.
  \newblock \emph{CoRR}, abs/2309.06126.
  
  \bibitem[{Ning et~al.(2024)Ning, Lin, Zhou, Wang, Yang, and Wang}]{DBLP:conf/iclr/Ning0ZWY024}
  Xuefei Ning, Zinan Lin, Zixuan Zhou, Zifu Wang, Huazhong Yang, and Yu~Wang. 2024.
  \newblock \href {https://openreview.net/forum?id=mqVgBbNCm9} {Skeleton-of-thought: Prompting llms for efficient parallel generation}.
  \newblock In \emph{The 12th International Conference on Learning Representations}.
  
  \bibitem[{Nye et~al.(2021)Nye, Andreassen, Gur{-}Ari, Michalewski, Austin, Bieber, Dohan, Lewkowycz, Bosma, Luan, Sutton, and Odena}]{DBLP:journals/corr/abs-2112-00114}
  Maxwell~I. Nye, Anders~Johan Andreassen, Guy Gur{-}Ari, Henryk Michalewski, Jacob Austin, David Bieber, David Dohan, Aitor Lewkowycz, Maarten Bosma, David Luan, Charles Sutton, and Augustus Odena. 2021.
  \newblock \href {https://arxiv.org/abs/2112.00114} {Show your work: Scratchpads for intermediate computation with language models}.
  \newblock \emph{CoRR}, abs/2112.00114.
  
  \bibitem[{OpenAI(2023)}]{DBLP:journals/corr/abs-2303-08774}
  OpenAI. 2023.
  \newblock \href {https://doi.org/10.48550/arXiv.2303.08774} {{GPT-4} technical report}.
  \newblock \emph{CoRR}, abs/2303.08774.
  
  \bibitem[{Ouyang et~al.(2022)Ouyang, Wu, Jiang, Almeida, Wainwright, Mishkin, Zhang, Agarwal, Slama, Ray, Schulman, Hilton, Kelton, Miller, Simens, Askell, Welinder, Christiano, Leike, and Lowe}]{DBLP:conf/nips/Ouyang0JAWMZASR22}
  Long Ouyang, Jeffrey Wu, Xu~Jiang, Diogo Almeida, Carroll~L. Wainwright, Pamela Mishkin, Chong Zhang, Sandhini Agarwal, Katarina Slama, Alex Ray, John Schulman, Jacob Hilton, Fraser Kelton, Luke Miller, Maddie Simens, Amanda Askell, Peter Welinder, Paul~F. Christiano, Jan Leike, and Ryan Lowe. 2022.
  \newblock \href {http://papers.nips.cc/paper\_files/paper/2022/hash/b1efde53be364a73914f58805a001731-Abstract-Conference.html} {Training language models to follow instructions with human feedback}.
  \newblock In \emph{Advances in Neural Information Processing Systems 35: Annual Conference on Neural Information Processing Systems 2022}.
  
  \bibitem[{Pesce and Montana(2023)}]{DBLP:journals/ml/PesceM23}
  Emanuele Pesce and Giovanni Montana. 2023.
  \newblock \href {https://doi.org/10.1007/s10994-022-06286-6} {Learning multi-agent coordination through connectivity-driven communication}.
  \newblock \emph{Mach. Learn.}, 112(2):483--514.
  
  \bibitem[{Plaat et~al.(2024)Plaat, Wong, Verberne, Broekens, van Stein, and B{\"{a}}ck}]{DBLP:journals/corr/abs-2407-11511}
  Aske Plaat, Annie Wong, Suzan Verberne, Joost Broekens, Niki van Stein, and Thomas B{\"{a}}ck. 2024.
  \newblock \href {https://doi.org/10.48550/ARXIV.2407.11511} {Reasoning with large language models, a survey}.
  \newblock \emph{CoRR}, abs/2407.11511.
  
  \bibitem[{Pope et~al.(2023)Pope, Douglas, Chowdhery, Devlin, Bradbury, Heek, Xiao, Agrawal, and Dean}]{DBLP:conf/mlsys/PopeDCDBHXAD23}
  Reiner Pope, Sholto Douglas, Aakanksha Chowdhery, Jacob Devlin, James Bradbury, Jonathan Heek, Kefan Xiao, Shivani Agrawal, and Jeff Dean. 2023.
  \newblock \href {https://proceedings.mlsys.org/paper\_files/paper/2023/hash/c4be71ab8d24cdfb45e3d06dbfca2780-Abstract-mlsys2023.html} {Efficiently scaling transformer inference}.
  \newblock In \emph{Proceedings of the Sixth Conference on Machine Learning and Systems}. mlsys.org.
  
  \bibitem[{Qian et~al.(2024{\natexlab{a}})Qian, Liu, Liu, Chen, Dang, Li, Yang, Chen, Su, Cong, Xu, Li, Liu, and Sun}]{DBLP:conf/acl/QianLLCDL0CSCXL24}
  Chen Qian, Wei Liu, Hongzhang Liu, Nuo Chen, Yufan Dang, Jiahao Li, Cheng Yang, Weize Chen, Yusheng Su, Xin Cong, Juyuan Xu, Dahai Li, Zhiyuan Liu, and Maosong Sun. 2024{\natexlab{a}}.
  \newblock \href {https://doi.org/10.18653/v1/2024.acl-long.810} {Chatdev: Communicative agents for software development}.
  \newblock In \emph{Proceedings of the 62nd Annual Meeting of the Association for Computational Linguistics}, pages 15174--15186.
  
  \bibitem[{Qian et~al.(2024{\natexlab{b}})Qian, Xie, Wang, Liu, Dang, Du, Chen, Yang, Liu, and Sun}]{DBLP:journals/corr/abs-2406-07155}
  Chen Qian, Zihao Xie, Yifei Wang, Wei Liu, Yufan Dang, Zhuoyun Du, Weize Chen, Cheng Yang, Zhiyuan Liu, and Maosong Sun. 2024{\natexlab{b}}.
  \newblock \href {https://doi.org/10.48550/arXiv.2406.07155} {Scaling large-language-model-based multi-agent collaboration}.
  \newblock \emph{CoRR}, abs/2406.07155.
  
  \bibitem[{Qiao et~al.(2024{\natexlab{a}})Qiao, Gui, Lv, Jia, Chen, and Zhang}]{DBLP:conf/naacl/QiaoGLJC024}
  Shuofei Qiao, Honghao Gui, Chengfei Lv, Qianghuai Jia, Huajun Chen, and Ningyu Zhang. 2024{\natexlab{a}}.
  \newblock \href {https://doi.org/10.18653/V1/2024.NAACL-LONG.195} {Making language models better tool learners with execution feedback}.
  \newblock In \emph{Proceedings of the 2024 Conference of the North American Chapter of the Association for Computational Linguistics: Human Language Technologies}, pages 3550--3568. Association for Computational Linguistics.
  
  \bibitem[{Qiao et~al.(2024{\natexlab{b}})Qiao, Zhang, Fang, Luo, Zhou, Jiang, Lv, and Chen}]{DBLP:conf/acl/Qiao0FLZJLC24}
  Shuofei Qiao, Ningyu Zhang, Runnan Fang, Yujie Luo, Wangchunshu Zhou, Yuchen~Eleanor Jiang, Chengfei Lv, and Huajun Chen. 2024{\natexlab{b}}.
  \newblock \href {https://doi.org/10.18653/v1/2024.acl-long.165} {Autoact: Automatic agent learning from scratch for {QA} via self-planning}.
  \newblock In \emph{Proceedings of the 62nd Annual Meeting of the Association for Computational Linguistics}, pages 3003--3021. Association for Computational Linguistics.
  
  \bibitem[{Rafailov et~al.(2023)Rafailov, Sharma, Mitchell, Manning, Ermon, and Finn}]{DBLP:conf/nips/RafailovSMMEF23}
  Rafael Rafailov, Archit Sharma, Eric Mitchell, Christopher~D. Manning, Stefano Ermon, and Chelsea Finn. 2023.
  \newblock \href {http://papers.nips.cc/paper\_files/paper/2023/hash/a85b405ed65c6477a4fe8302b5e06ce7-Abstract-Conference.html} {Direct preference optimization: Your language model is secretly a reward model}.
  \newblock In \emph{Advances in Neural Information Processing Systems 36: Annual Conference on Neural Information Processing Systems 2023}.
  
  \bibitem[{Rozi{\`{e}}re et~al.(2023)Rozi{\`{e}}re, Gehring, Gloeckle, Sootla, Gat, Tan, Adi, Liu, Remez, Rapin, Kozhevnikov, Evtimov, Bitton, Bhatt, Canton{-}Ferrer, Grattafiori, Xiong, D{\'{e}}fossez, Copet, Azhar, Touvron, Martin, Usunier, Scialom, and Synnaeve}]{DBLP:journals/corr/abs-2308-12950}
  Baptiste Rozi{\`{e}}re, Jonas Gehring, Fabian Gloeckle, Sten Sootla, Itai Gat, Xiaoqing~Ellen Tan, Yossi Adi, Jingyu Liu, Tal Remez, J{\'{e}}r{\'{e}}my Rapin, Artyom Kozhevnikov, Ivan Evtimov, Joanna Bitton, Manish Bhatt, Cristian Canton{-}Ferrer, Aaron Grattafiori, Wenhan Xiong, Alexandre D{\'{e}}fossez, Jade Copet, Faisal Azhar, Hugo Touvron, Louis Martin, Nicolas Usunier, Thomas Scialom, and Gabriel Synnaeve. 2023.
  \newblock \href {https://doi.org/10.48550/arXiv.2308.12950} {Code llama: Open foundation models for code}.
  \newblock \emph{CoRR}, abs/2308.12950.
  
  \bibitem[{Schick et~al.(2023)Schick, Dwivedi{-}Yu, Dess{\`{\i}}, Raileanu, Lomeli, Hambro, Zettlemoyer, Cancedda, and Scialom}]{DBLP:conf/nips/SchickDDRLHZCS23}
  Timo Schick, Jane Dwivedi{-}Yu, Roberto Dess{\`{\i}}, Roberta Raileanu, Maria Lomeli, Eric Hambro, Luke Zettlemoyer, Nicola Cancedda, and Thomas Scialom. 2023.
  \newblock \href {http://papers.nips.cc/paper\_files/paper/2023/hash/d842425e4bf79ba039352da0f658a906-Abstract-Conference.html} {Toolformer: Language models can teach themselves to use tools}.
  \newblock In \emph{Advances in Neural Information Processing Systems 36: Annual Conference on Neural Information Processing Systems 2023}.
  
  \bibitem[{Schulman et~al.(2017)Schulman, Wolski, Dhariwal, Radford, and Klimov}]{DBLP:journals/corr/SchulmanWDRK17}
  John Schulman, Filip Wolski, Prafulla Dhariwal, Alec Radford, and Oleg Klimov. 2017.
  \newblock \href {http://arxiv.org/abs/1707.06347} {Proximal policy optimization algorithms}.
  \newblock \emph{CoRR}, abs/1707.06347.
  
  \bibitem[{Sel et~al.(2024)Sel, Al{-}Tawaha, Khattar, Jia, and Jin}]{DBLP:conf/icml/SelAK0024}
  Bilgehan Sel, Ahmad Al{-}Tawaha, Vanshaj Khattar, Ruoxi Jia, and Ming Jin. 2024.
  \newblock \href {https://openreview.net/forum?id=KJL2b6BthC} {Algorithm of thoughts: Enhancing exploration of ideas in large language models}.
  \newblock In \emph{Forty-first International Conference on Machine Learning}.
  
  \bibitem[{Setlur et~al.(2024)Setlur, Nagpal, Fisch, Geng, Eisenstein, Agarwal, Agarwal, Berant, and Kumar}]{DBLP:journals/corr/abs-2410-08146}
  Amrith Setlur, Chirag Nagpal, Adam Fisch, Xinyang Geng, Jacob Eisenstein, Rishabh Agarwal, Alekh Agarwal, Jonathan Berant, and Aviral Kumar. 2024.
  \newblock \href {https://doi.org/10.48550/ARXIV.2410.08146} {Rewarding progress: Scaling automated process verifiers for {LLM} reasoning}.
  \newblock \emph{CoRR}, abs/2410.08146.
  
  \bibitem[{Shao et~al.(2024)Shao, Wang, Zhu, Xu, Song, Zhang, Li, Wu, and Guo}]{DBLP:journals/corr/abs-2402-03300}
  Zhihong Shao, Peiyi Wang, Qihao Zhu, Runxin Xu, Junxiao Song, Mingchuan Zhang, Y.~K. Li, Y.~Wu, and Daya Guo. 2024.
  \newblock \href {https://doi.org/10.48550/ARXIV.2402.03300} {Deepseekmath: Pushing the limits of mathematical reasoning in open language models}.
  \newblock \emph{CoRR}, abs/2402.03300.
  
  \bibitem[{Sheng et~al.(2024)Sheng, Zhang, Ye, Wu, Zhang, Zhang, Peng, Lin, and Wu}]{DBLP:journals/corr/abs-2409-19256}
  Guangming Sheng, Chi Zhang, Zilingfeng Ye, Xibin Wu, Wang Zhang, Ru~Zhang, Yanghua Peng, Haibin Lin, and Chuan Wu. 2024.
  \newblock \href {https://doi.org/10.48550/ARXIV.2409.19256} {Hybridflow: {A} flexible and efficient {RLHF} framework}.
  \newblock \emph{CoRR}, abs/2409.19256.
  
  \bibitem[{Shinn et~al.(2023)Shinn, Labash, and Gopinath}]{DBLP:journals/corr/abs-2303-11366}
  Noah Shinn, Beck Labash, and Ashwin Gopinath. 2023.
  \newblock \href {https://doi.org/10.48550/arXiv.2303.11366} {Reflexion: an autonomous agent with dynamic memory and self-reflection}.
  \newblock \emph{CoRR}, abs/2303.11366.
  
  \bibitem[{Shridhar et~al.(2023)Shridhar, Stolfo, and Sachan}]{DBLP:conf/acl/ShridharSS23}
  Kumar Shridhar, Alessandro Stolfo, and Mrinmaya Sachan. 2023.
  \newblock \href {https://doi.org/10.18653/V1/2023.FINDINGS-ACL.441} {Distilling reasoning capabilities into smaller language models}.
  \newblock In \emph{Findings of the Association for Computational Linguistics: {ACL} 2023}, pages 7059--7073. Association for Computational Linguistics.
  
  \bibitem[{Snell et~al.(2024)Snell, Lee, Xu, and Kumar}]{DBLP:journals/corr/abs-2408-03314}
  Charlie Snell, Jaehoon Lee, Kelvin Xu, and Aviral Kumar. 2024.
  \newblock \href {https://doi.org/10.48550/arXiv.2408.03314} {Scaling {LLM} test-time compute optimally can be more effective than scaling model parameters}.
  \newblock \emph{CoRR}, abs/2408.03314.
  
  \bibitem[{Song et~al.(2024)Song, Yin, Yue, Huang, Li, and Lin}]{DBLP:journals/corr/abs-2403-02502}
  Yifan Song, Da~Yin, Xiang Yue, Jie Huang, Sujian Li, and Bill~Yuchen Lin. 2024.
  \newblock \href {https://doi.org/10.48550/ARXIV.2403.02502} {Trial and error: Exploration-based trajectory optimization for {LLM} agents}.
  \newblock \emph{CoRR}, abs/2403.02502.
  
  \bibitem[{Tang et~al.(2024)Tang, Zhang, Wang, and Wei}]{DBLP:conf/icml/TangZWW24}
  Zhengyang Tang, Xingxing Zhang, Benyou Wang, and Furu Wei. 2024.
  \newblock \href {https://openreview.net/forum?id=Kjww7ZN47M} {Mathscale: Scaling instruction tuning for mathematical reasoning}.
  \newblock In \emph{Forty-first International Conference on Machine Learning}. OpenReview.net.
  
  \bibitem[{Touvron et~al.(2023)Touvron, Martin, Stone, Albert, Almahairi, Babaei, Bashlykov, Batra, Bhargava, Bhosale, Bikel, Blecher, Canton{-}Ferrer, Chen, Cucurull, Esiobu, Fernandes, Fu, Fu, Fuller, Gao, Goswami, Goyal, Hartshorn, Hosseini, Hou, Inan, Kardas, Kerkez, Khabsa, Kloumann, Korenev, Koura, Lachaux, Lavril, Lee, Liskovich, Lu, Mao, Martinet, Mihaylov, Mishra, Molybog, Nie, Poulton, Reizenstein, Rungta, Saladi, Schelten, Silva, Smith, Subramanian, Tan, Tang, Taylor, Williams, Kuan, Xu, Yan, Zarov, Zhang, Fan, Kambadur, Narang, Rodriguez, Stojnic, Edunov, and Scialom}]{DBLP:journals/corr/abs-2307-09288}
  Hugo Touvron, Louis Martin, Kevin Stone, Peter Albert, Amjad Almahairi, Yasmine Babaei, Nikolay Bashlykov, Soumya Batra, Prajjwal Bhargava, Shruti Bhosale, Dan Bikel, Lukas Blecher, Cristian Canton{-}Ferrer, Moya Chen, Guillem Cucurull, David Esiobu, Jude Fernandes, Jeremy Fu, Wenyin Fu, Brian Fuller, Cynthia Gao, Vedanuj Goswami, Naman Goyal, Anthony Hartshorn, Saghar Hosseini, Rui Hou, Hakan Inan, Marcin Kardas, Viktor Kerkez, Madian Khabsa, Isabel Kloumann, Artem Korenev, Punit~Singh Koura, Marie{-}Anne Lachaux, Thibaut Lavril, Jenya Lee, Diana Liskovich, Yinghai Lu, Yuning Mao, Xavier Martinet, Todor Mihaylov, Pushkar Mishra, Igor Molybog, Yixin Nie, Andrew Poulton, Jeremy Reizenstein, Rashi Rungta, Kalyan Saladi, Alan Schelten, Ruan Silva, Eric~Michael Smith, Ranjan Subramanian, Xiaoqing~Ellen Tan, Binh Tang, Ross Taylor, Adina Williams, Jian~Xiang Kuan, Puxin Xu, Zheng Yan, Iliyan Zarov, Yuchen Zhang, Angela Fan, Melanie Kambadur, Sharan Narang, Aur{\'{e}}lien Rodriguez, Robert Stojnic, Sergey Edunov,
    and Thomas Scialom. 2023.
  \newblock \href {https://doi.org/10.48550/arXiv.2307.09288} {Llama 2: Open foundation and fine-tuned chat models}.
  \newblock \emph{CoRR}, abs/2307.09288.
  
  \bibitem[{Trung et~al.(2024)Trung, Zhang, Jie, Sun, Jin, and Li}]{DBLP:conf/acl/TrungZJSJL24}
  Luong~Quoc Trung, Xinbo Zhang, Zhanming Jie, Peng Sun, Xiaoran Jin, and Hang Li. 2024.
  \newblock \href {https://doi.org/10.18653/V1/2024.ACL-LONG.410} {Reft: Reasoning with reinforced fine-tuning}.
  \newblock In \emph{Proceedings of the 62nd Annual Meeting of the Association for Computational Linguistics}, pages 7601--7614. Association for Computational Linguistics.
  
  \bibitem[{Wang et~al.(2024{\natexlab{a}})Wang, Deng, Lv, Liang, He, Yan, and An}]{DBLP:journals/corr/abs-2406-14283}
  Chaojie Wang, Yanchen Deng, Zhiyi Lv, Zeng Liang, Jujie He, Shuicheng Yan, and Bo~An. 2024{\natexlab{a}}.
  \newblock \href {https://doi.org/10.48550/ARXIV.2406.14283} {Q*: Improving multi-step reasoning for llms with deliberative planning}.
  \newblock \emph{CoRR}, abs/2406.14283.
  
  \bibitem[{Wang et~al.(2023{\natexlab{a}})Wang, Yan, Zhang, and Huang}]{DBLP:journals/corr/abs-2311-13126}
  Chengyu Wang, Junbing Yan, Wei Zhang, and Jun Huang. 2023{\natexlab{a}}.
  \newblock \href {https://doi.org/10.48550/ARXIV.2311.13126} {Towards better parameter-efficient fine-tuning for large language models: {A} position paper}.
  \newblock \emph{CoRR}, abs/2311.13126.
  
  \bibitem[{Wang et~al.(2024{\natexlab{b}})Wang, Qin, Lin, Pan, and Wong}]{DBLP:conf/sigir/0003QLPW24}
  Hongru Wang, Yujia Qin, Yankai Lin, Jeff~Z. Pan, and Kam{-}Fai Wong. 2024{\natexlab{b}}.
  \newblock \href {https://doi.org/10.1145/3626772.3661381} {Empowering large language models: Tool learning for real-world interaction}.
  \newblock In \emph{Proceedings of the 47th International {ACM} {SIGIR} Conference on Research and Development in Information Retrieval}, pages 2983--2986. {ACM}.
  
  \bibitem[{Wang et~al.(2024{\natexlab{c}})Wang, Li, Shao, Xu, Dai, Li, Chen, Wu, and Sui}]{DBLP:conf/acl/WangLSXDLCWS24}
  Peiyi Wang, Lei Li, Zhihong Shao, Runxin Xu, Damai Dai, Yifei Li, Deli Chen, Yu~Wu, and Zhifang Sui. 2024{\natexlab{c}}.
  \newblock \href {https://doi.org/10.18653/V1/2024.ACL-LONG.510} {Math-shepherd: Verify and reinforce llms step-by-step without human annotations}.
  \newblock In \emph{Proceedings of the 62nd Annual Meeting of the Association for Computational Linguistics}, pages 9426--9439. Association for Computational Linguistics.
  
  \bibitem[{Wang et~al.(2024{\natexlab{d}})Wang, Chen, Han, and Bai}]{DBLP:journals/corr/abs-2409-08642}
  Tianlong Wang, Junzhe Chen, Xueting Han, and Jing Bai. 2024{\natexlab{d}}.
  \newblock \href {https://doi.org/10.48550/ARXIV.2409.08642} {{CPL:} critical plan step learning boosts {LLM} generalization in reasoning tasks}.
  \newblock \emph{CoRR}, abs/2409.08642.
  
  \bibitem[{Wang et~al.(2024{\natexlab{e}})Wang, Kim, Rahman, Mitra, and Miao}]{DBLP:conf/chi/Wang0RMM24}
  Xinru Wang, Hannah Kim, Sajjadur Rahman, Kushan Mitra, and Zhengjie Miao. 2024{\natexlab{e}}.
  \newblock \href {https://doi.org/10.1145/3613904.3641960} {Human-llm collaborative annotation through effective verification of {LLM} labels}.
  \newblock In \emph{Proceedings of the {CHI} Conference on Human Factors in Computing Systems}, pages 303:1--303:21. {ACM}.
  
  \bibitem[{Wang et~al.(2023{\natexlab{b}})Wang, Wei, Schuurmans, Le, Chi, Narang, Chowdhery, and Zhou}]{DBLP:conf/iclr/0002WSLCNCZ23}
  Xuezhi Wang, Jason Wei, Dale Schuurmans, Quoc~V. Le, Ed~H. Chi, Sharan Narang, Aakanksha Chowdhery, and Denny Zhou. 2023{\natexlab{b}}.
  \newblock \href {https://openreview.net/forum?id=1PL1NIMMrw} {Self-consistency improves chain of thought reasoning in language models}.
  \newblock In \emph{The Eleventh International Conference on Learning Representations}.
  
  \bibitem[{Wang et~al.(2024{\natexlab{f}})Wang, Mao, Wu, Ge, Wei, and Ji}]{DBLP:conf/naacl/WangMW0WJ24}
  Zhenhailong Wang, Shaoguang Mao, Wenshan Wu, Tao Ge, Furu Wei, and Heng Ji. 2024{\natexlab{f}}.
  \newblock \href {https://doi.org/10.18653/v1/2024.naacl-long.15} {Unleashing the emergent cognitive synergy in large language models: {A} task-solving agent through multi-persona self-collaboration}.
  \newblock In \emph{Proceedings of the 2024 Conference of the North American Chapter of the Association for Computational Linguistics: Human Language Technologies}, pages 257--279.
  
  \bibitem[{Wang et~al.(2025)Wang, Wang, Liu, Ding, Zhang, Liu, and Zhang}]{agentdropout}
  Zhexuan Wang, Yutong Wang, Xuebo Liu, Liang Ding, Miao Zhang, Jie Liu, and Min Zhang. 2025.
  \newblock \href {https://arxiv.org/abs/2503.18891} {Agentdropout: Dynamic agent elimination for token-efficient and high-performance llm-based multi-agent collaboration}.
  \newblock \emph{CoRR}, abs/2503.18891.
  
  \bibitem[{Wang et~al.(2024{\natexlab{g}})Wang, Li, Wu, Luo, Hou, Yu, and Shang}]{DBLP:conf/emnlp/WangLWLH0S24}
  Zihan Wang, Yunxuan Li, Yuexin Wu, Liangchen Luo, Le~Hou, Hongkun Yu, and Jingbo Shang. 2024{\natexlab{g}}.
  \newblock \href {https://aclanthology.org/2024.findings-emnlp.429} {Multi-step problem solving through a verifier: An empirical analysis on model-induced process supervision}.
  \newblock In \emph{Findings of the Association for Computational Linguistics: {EMNLP} 2024}, pages 7309--7319. Association for Computational Linguistics.
  
  \bibitem[{Wei et~al.(2022)Wei, Wang, Schuurmans, Bosma, Ichter, Xia, Chi, Le, and Zhou}]{DBLP:conf/nips/Wei0SBIXCLZ22}
  Jason Wei, Xuezhi Wang, Dale Schuurmans, Maarten Bosma, Brian Ichter, Fei Xia, Ed~H. Chi, Quoc~V. Le, and Denny Zhou. 2022.
  \newblock \href {http://papers.nips.cc/paper\_files/paper/2022/hash/9d5609613524ecf4f15af0f7b31abca4-Abstract-Conference.html} {Chain-of-thought prompting elicits reasoning in large language models}.
  \newblock In \emph{Advances in Neural Information Processing Systems 35: Annual Conference on Neural Information Processing Systems 2022}.
  
  \bibitem[{Welleck et~al.(2023)Welleck, Lu, West, Brahman, Shen, Khashabi, and Choi}]{DBLP:conf/iclr/WelleckLWBSK023}
  Sean Welleck, Ximing Lu, Peter West, Faeze Brahman, Tianxiao Shen, Daniel Khashabi, and Yejin Choi. 2023.
  \newblock \href {https://openreview.net/forum?id=hH36JeQZDaO} {Generating sequences by learning to self-correct}.
  \newblock In \emph{The Eleventh International Conference on Learning Representations}. OpenReview.net.
  
  \bibitem[{Wen et~al.(2025)Wen, Cai, Xiao, He, An, Duan, Du, Liu, Tang, Lv, Zou, Deng, Jia, and Zhang}]{wen2025light}
  Liang Wen, Yunke Cai, Fenrui Xiao, Xin He, Qi~An, Zhenyu Duan, Yimin Du, Junchen Liu, Lifu Tang, Xiaowei Lv, Haosheng Zou, Yongchao Deng, Shousheng Jia, and Xiangzheng Zhang. 2025.
  \newblock Light-r1: Curriculum sft, dpo and rl for long cot from scratch and beyond.
  \newblock \emph{arXiv preprint arXiv:2503.10460}.
  
  \bibitem[{Wu et~al.(2023{\natexlab{a}})Wu, Bansal, Zhang, Wu, Zhang, Zhu, Li, Jiang, Zhang, and Wang}]{DBLP:journals/corr/abs-2308-08155}
  Qingyun Wu, Gagan Bansal, Jieyu Zhang, Yiran Wu, Shaokun Zhang, Erkang Zhu, Beibin Li, Li~Jiang, Xiaoyun Zhang, and Chi Wang. 2023{\natexlab{a}}.
  \newblock \href {https://doi.org/10.48550/arXiv.2308.08155} {Autogen: Enabling next-gen {LLM} applications via multi-agent conversation framework}.
  \newblock \emph{CoRR}, abs/2308.08155.
  
  \bibitem[{Wu et~al.(2023{\natexlab{b}})Wu, Irsoy, Lu, Dabravolski, Dredze, Gehrmann, Kambadur, Rosenberg, and Mann}]{DBLP:journals/corr/abs-2303-17564}
  Shijie Wu, Ozan Irsoy, Steven Lu, Vadim Dabravolski, Mark Dredze, Sebastian Gehrmann, Prabhanjan Kambadur, David~S. Rosenberg, and Gideon Mann. 2023{\natexlab{b}}.
  \newblock \href {https://doi.org/10.48550/arXiv.2303.17564} {Bloomberggpt: {A} large language model for finance}.
  \newblock \emph{CoRR}, abs/2303.17564.
  
  \bibitem[{Wu et~al.(2024)Wu, Sun, Li, Welleck, and Yang}]{DBLP:journals/corr/abs-2408-00724}
  Yangzhen Wu, Zhiqing Sun, Shanda Li, Sean Welleck, and Yiming Yang. 2024.
  \newblock \href {https://doi.org/10.48550/arXiv.2408.00724} {An empirical analysis of compute-optimal inference for problem-solving with language models}.
  \newblock \emph{CoRR}, abs/2408.00724.
  
  \bibitem[{Xie et~al.(2024)Xie, Goyal, Zheng, Kan, Lillicrap, Kawaguchi, and Shieh}]{DBLP:journals/corr/abs-2405-00451}
  Yuxi Xie, Anirudh Goyal, Wenyue Zheng, Min{-}Yen Kan, Timothy~P. Lillicrap, Kenji Kawaguchi, and Michael Shieh. 2024.
  \newblock \href {https://doi.org/10.48550/ARXIV.2405.00451} {Monte carlo tree search boosts reasoning via iterative preference learning}.
  \newblock \emph{CoRR}, abs/2405.00451.
  
  \bibitem[{Xie et~al.(2023)Xie, Kawaguchi, Zhao, Zhao, Kan, He, and Xie}]{DBLP:conf/nips/XieKZZKHX23}
  Yuxi Xie, Kenji Kawaguchi, Yiran Zhao, James~Xu Zhao, Min{-}Yen Kan, Junxian He, and Michael~Qizhe Xie. 2023.
  \newblock \href {http://papers.nips.cc/paper\_files/paper/2023/hash/81fde95c4dc79188a69ce5b24d63010b-Abstract-Conference.html} {Self-evaluation guided beam search for reasoning}.
  \newblock In \emph{Advances in Neural Information Processing Systems 36: Annual Conference on Neural Information Processing Systems 2023}.
  
  \bibitem[{Xin et~al.(2024)Xin, Guo, Shao, Ren, Zhu, Liu, Ruan, Li, and Liang}]{DBLP:journals/corr/abs-2405-14333}
  Huajian Xin, Daya Guo, Zhihong Shao, Zhizhou Ren, Qihao Zhu, Bo~Liu, Chong Ruan, Wenda Li, and Xiaodan Liang. 2024.
  \newblock \href {https://doi.org/10.48550/arXiv.2405.14333} {Deepseek-prover: Advancing theorem proving in llms through large-scale synthetic data}.
  \newblock \emph{CoRR}, abs/2405.14333.
  
  \bibitem[{Xu et~al.(2025)Xu, Hao, Zong, Wang, Zhang, Wang, Lan, Gong, Ouyang, Meng, Shao, Yan, Yang, Song, Ren, Hu, Li, Feng, Gao, and Li}]{DBLP:journals/corr/abs-2501-09686}
  Fengli Xu, Qianyue Hao, Zefang Zong, Jingwei Wang, Yunke Zhang, Jingyi Wang, Xiaochong Lan, Jiahui Gong, Tianjian Ouyang, Fanjin Meng, Chenyang Shao, Yuwei Yan, Qinglong Yang, Yiwen Song, Sijian Ren, Xinyuan Hu, Yu~Li, Jie Feng, Chen Gao, and Yong Li. 2025.
  \newblock \href {https://doi.org/10.48550/ARXIV.2501.09686} {Towards large reasoning models: {A} survey of reinforced reasoning with large language models}.
  \newblock \emph{CoRR}, abs/2501.09686.
  
  \bibitem[{Yan et~al.(2023)Yan, Wang, Zhang, He, Huang, and Zhang}]{DBLP:conf/emnlp/Yan0ZHHZ23}
  Junbing Yan, Chengyu Wang, Taolin Zhang, Xiaofeng He, Jun Huang, and Wei Zhang. 2023.
  \newblock \href {https://doi.org/10.18653/V1/2023.FINDINGS-EMNLP.828} {From complex to simple: Unraveling the cognitive tree for reasoning with small language models}.
  \newblock In \emph{Findings of the Association for Computational Linguistics: {EMNLP} 2023}, pages 12413--12425. Association for Computational Linguistics.
  
  \bibitem[{Yan et~al.(2024)Yan, Zhang, and Huang}]{DBLP:journals/corr/abs-2403-17674}
  Yikuan Yan, Yaolun Zhang, and Keman Huang. 2024.
  \newblock \href {https://doi.org/10.48550/arXiv.2403.17674} {Depending on yourself when you should: Mentoring {LLM} with {RL} agents to become the master in cybersecurity games}.
  \newblock \emph{CoRR}, abs/2403.17674.
  
  \bibitem[{Yang et~al.(2024{\natexlab{a}})Yang, Zhang, Hui, Gao, Yu, Li, Liu, Tu, Zhou, Lin, Lu, Xue, Lin, Liu, Ren, and Zhang}]{DBLP:journals/corr/abs-2409-12122}
  An~Yang, Beichen Zhang, Binyuan Hui, Bofei Gao, Bowen Yu, Chengpeng Li, Dayiheng Liu, Jianhong Tu, Jingren Zhou, Junyang Lin, Keming Lu, Mingfeng Xue, Runji Lin, Tianyu Liu, Xingzhang Ren, and Zhenru Zhang. 2024{\natexlab{a}}.
  \newblock \href {https://doi.org/10.48550/ARXIV.2409.12122} {Qwen2.5-math technical report: Toward mathematical expert model via self-improvement}.
  \newblock \emph{CoRR}, abs/2409.12122.
  
  \bibitem[{Yang et~al.(2024{\natexlab{b}})Yang, Zhang, Kuang, Xie, Huang, and Ananiadou}]{DBLP:conf/www/YangZKXHA24}
  Kailai Yang, Tianlin Zhang, Ziyan Kuang, Qianqian Xie, Jimin Huang, and Sophia Ananiadou. 2024{\natexlab{b}}.
  \newblock \href {https://doi.org/10.1145/3589334.3648137} {Mentallama: Interpretable mental health analysis on social media with large language models}.
  \newblock In \emph{Proceedings of the {ACM} on Web Conference 2024}, pages 4489--4500.
  
  \bibitem[{Yang et~al.(2023)Yang, Sun, Li, Liu, Li, Liu, Huang, and Gao}]{DBLP:journals/corr/abs-2310-15777}
  Yizhe Yang, Huashan Sun, Jiawei Li, Runheng Liu, Yinghao Li, Yuhang Liu, Heyan Huang, and Yang Gao. 2023.
  \newblock \href {https://doi.org/10.48550/arXiv.2310.15777} {Mindllm: Pre-training lightweight large language model from scratch, evaluations and domain applications}.
  \newblock \emph{CoRR}, abs/2310.15777.
  
  \bibitem[{Yang et~al.(2024{\natexlab{c}})Yang, Pang, Feng, Wang, Chen, Zhu, and Liu}]{DBLP:conf/acl/YangPFWCZL24}
  Zhaorui Yang, Tianyu Pang, Haozhe Feng, Han Wang, Wei Chen, Minfeng Zhu, and Qian Liu. 2024{\natexlab{c}}.
  \newblock \href {https://doi.org/10.18653/V1/2024.ACL-LONG.58} {Self-distillation bridges distribution gap in language model fine-tuning}.
  \newblock In \emph{Proceedings of the 62nd Annual Meeting of the Association for Computational Linguistics}, pages 1028--1043. Association for Computational Linguistics.
  
  \bibitem[{Yao et~al.(2023)Yao, Yu, Zhao, Shafran, Griffiths, Cao, and Narasimhan}]{DBLP:conf/nips/YaoYZS00N23}
  Shunyu Yao, Dian Yu, Jeffrey Zhao, Izhak Shafran, Tom Griffiths, Yuan Cao, and Karthik Narasimhan. 2023.
  \newblock \href {http://papers.nips.cc/paper\_files/paper/2023/hash/271db9922b8d1f4dd7aaef84ed5ac703-Abstract-Conference.html} {Tree of thoughts: Deliberate problem solving with large language models}.
  \newblock In \emph{Advances in Neural Information Processing Systems 36: Annual Conference on Neural Information Processing Systems 2023}.
  
  \bibitem[{Yao et~al.(2021)Yao, Huang, Wang, Dong, and Wei}]{DBLP:conf/acl/YaoHWDW21}
  Yunzhi Yao, Shaohan Huang, Wenhui Wang, Li~Dong, and Furu Wei. 2021.
  \newblock \href {https://doi.org/10.18653/v1/2021.findings-acl.40} {Adapt-and-distill: Developing small, fast and effective pretrained language models for domains}.
  \newblock In \emph{Findings of the Association for Computational Linguistics: {ACL/IJCNLP} 2021}, pages 460--470.
  
  \bibitem[{Ying et~al.(2024)Ying, Zhang, Li, Zhou, Shao, Fei, Ma, Hong, Liu, Wang, Wang, Wu, Li, Zhou, Liu, Zhang, Zhang, Yan, Qiu, Wang, Chen, and Lin}]{DBLP:journals/corr/abs-2402-06332}
  Huaiyuan Ying, Shuo Zhang, Linyang Li, Zhejian Zhou, Yunfan Shao, Zhaoye Fei, Yichuan Ma, Jiawei Hong, Kuikun Liu, Ziyi Wang, Yudong Wang, Zijian Wu, Shuaibin Li, Fengzhe Zhou, Hongwei Liu, Songyang Zhang, Wenwei Zhang, Hang Yan, Xipeng Qiu, Jiayu Wang, Kai Chen, and Dahua Lin. 2024.
  \newblock \href {https://doi.org/10.48550/ARXIV.2402.06332} {Internlm-math: Open math large language models toward verifiable reasoning}.
  \newblock \emph{CoRR}, abs/2402.06332.
  
  \bibitem[{Yuan et~al.(2021)Yuan, Zhao, Du, Ding, Liu, Cen, Zou, Yang, and Tang}]{DBLP:journals/aiopen/YuanZDDLCZYT21}
  Sha Yuan, Hanyu Zhao, Zhengxiao Du, Ming Ding, Xiao Liu, Yukuo Cen, Xu~Zou, Zhilin Yang, and Jie Tang. 2021.
  \newblock \href {https://doi.org/10.1016/j.aiopen.2021.06.001} {Wudaocorpora: {A} super large-scale chinese corpora for pre-training language models}.
  \newblock \emph{{AI} Open}, 2:65--68.
  
  \bibitem[{Yuan et~al.(2025)Yuan, Yu, Jiang, Padthe, Li, Wang, Kulikov, Cho, Tian, Weston, and Li}]{yuan2025naturalreasoningreasoningwild28m}
  Weizhe Yuan, Jane Yu, Song Jiang, Karthik Padthe, Yang Li, Dong Wang, Ilia Kulikov, Kyunghyun Cho, Yuandong Tian, Jason~E Weston, and Xian Li. 2025.
  \newblock \href {https://arxiv.org/abs/2502.13124} {Naturalreasoning: Reasoning in the wild with 2.8m challenging questions}.
  \newblock \emph{CoRR}, abs/2502.13124.
  
  \bibitem[{Yue et~al.(2023)Yue, Chen, Wang, Li, Shen, Liu, Zhou, Xiao, Yun, Huang, and Wei}]{DBLP:journals/corr/abs-2309-11325}
  Shengbin Yue, Wei Chen, Siyuan Wang, Bingxuan Li, Chenchen Shen, Shujun Liu, Yuxuan Zhou, Yao Xiao, Song Yun, Xuanjing Huang, and Zhongyu Wei. 2023.
  \newblock \href {https://doi.org/10.48550/arXiv.2309.11325} {Disc-lawllm: Fine-tuning large language models for intelligent legal services}.
  \newblock \emph{CoRR}, abs/2309.11325.
  
  \bibitem[{Yue et~al.(2024{\natexlab{a}})Yue, Wang, Huang, and Wang}]{DBLP:journals/corr/abs-2412-04871}
  Yuanhao Yue, Chengyu Wang, Jun Huang, and Peng Wang. 2024{\natexlab{a}}.
  \newblock \href {https://doi.org/10.48550/ARXIV.2412.04871} {Building a family of data augmentation models for low-cost {LLM} fine-tuning on the cloud}.
  \newblock \emph{CoRR}, abs/2412.04871.
  
  \bibitem[{Yue et~al.(2024{\natexlab{b}})Yue, Wang, Huang, and Wang}]{DBLP:conf/emnlp/YueWHW24}
  Yuanhao Yue, Chengyu Wang, Jun Huang, and Peng Wang. 2024{\natexlab{b}}.
  \newblock \href {https://aclanthology.org/2024.findings-emnlp.350} {Distilling instruction-following abilities of large language models with task-aware curriculum planning}.
  \newblock In \emph{Findings of the Association for Computational Linguistics: {EMNLP} 2024}, pages 6030--6054. Association for Computational Linguistics.
  
  \bibitem[{Zelikman et~al.(2023)Zelikman, Lorch, Mackey, and Kalai}]{DBLP:journals/corr/abs-2310-02304}
  Eric Zelikman, Eliana Lorch, Lester Mackey, and Adam~Tauman Kalai. 2023.
  \newblock \href {https://doi.org/10.48550/arXiv.2310.02304} {Self-taught optimizer {(STOP):} recursively self-improving code generation}.
  \newblock \emph{CoRR}, abs/2310.02304.
  
  \bibitem[{Zhang et~al.(2024{\natexlab{a}})Zhang, Liu, Cherry, and Firat}]{DBLP:conf/iclr/0006LCF24}
  Biao Zhang, Zhongtao Liu, Colin Cherry, and Orhan Firat. 2024{\natexlab{a}}.
  \newblock \href {https://openreview.net/forum?id=5HCnKDeTws} {When scaling meets {LLM} finetuning: The effect of data, model and finetuning method}.
  \newblock In \emph{The Twelfth International Conference on Learning Representations}. OpenReview.net.
  
  \bibitem[{Zhang et~al.(2024{\natexlab{b}})Zhang, Hu, Zhoubian, Du, Yang, Wang, Yue, Dong, and Tang}]{DBLP:journals/corr/abs-2401-07950}
  Dan Zhang, Ziniu Hu, Sining Zhoubian, Zhengxiao Du, Kaiyu Yang, Zihan Wang, Yisong Yue, Yuxiao Dong, and Jie Tang. 2024{\natexlab{b}}.
  \newblock \href {https://doi.org/10.48550/arXiv.2401.07950} {Sciglm: Training scientific language models with self-reflective instruction annotation and tuning}.
  \newblock \emph{CoRR}, abs/2401.07950.
  
  \bibitem[{Zhang et~al.(2024{\natexlab{c}})Zhang, Zhoubian, Hu, Yue, Dong, and Tang}]{DBLP:conf/nips/ZhangZHYD024}
  Dan Zhang, Sining Zhoubian, Ziniu Hu, Yisong Yue, Yuxiao Dong, and Jie Tang. 2024{\natexlab{c}}.
  \newblock \href {http://papers.nips.cc/paper\_files/paper/2024/hash/76ec4dc30e9faaf0e4b6093eaa377218-Abstract-Conference.html} {Rest-mcts*: {LLM} self-training via process reward guided tree search}.
  \newblock In \emph{Advances in Neural Information Processing Systems 38: Annual Conference on Neural Information Processing Systems 2024}.
  
  \bibitem[{Zhang et~al.(2024{\natexlab{d}})Zhang, Liu, Tan, Chen, Yan, Yan, Li, Huang, Yue, Zhou, Zhang, Su, Zhong, Li, and Ouyang}]{DBLP:journals/corr/abs-2402-06852}
  Di~Zhang, Wei Liu, Qian Tan, Jingdan Chen, Hang Yan, Yuliang Yan, Jiatong Li, Weiran Huang, Xiangyu Yue, Dongzhan Zhou, Shufei Zhang, Mao Su, Hansen Zhong, Yuqiang Li, and Wanli Ouyang. 2024{\natexlab{d}}.
  \newblock \href {https://doi.org/10.48550/arXiv.2402.06852} {Chemllm: {A} chemical large language model}.
  \newblock \emph{CoRR}, abs/2402.06852.
  
  \bibitem[{Zhang et~al.(2024{\natexlab{e}})Zhang, Yue, Li, Yun, Wan, Wang, Cheng, Yu, and Chen}]{DBLP:journals/corr/abs-2410-02506}
  Guibin Zhang, Yanwei Yue, Zhixun Li, Sukwon Yun, Guancheng Wan, Kun Wang, Dawei Cheng, Jeffrey~Xu Yu, and Tianlong Chen. 2024{\natexlab{e}}.
  \newblock \href {https://doi.org/10.48550/arXiv.2410.02506} {Cut the crap: An economical communication pipeline for llm-based multi-agent systems}.
  \newblock \emph{CoRR}, abs/2410.02506.
  
  \bibitem[{Zhang et~al.(2024{\natexlab{f}})Zhang, Xu, Zhang, Liu, Hooi, and Deng}]{DBLP:conf/acl/ZhangX0LHD24}
  Jintian Zhang, Xin Xu, Ningyu Zhang, Ruibo Liu, Bryan Hooi, and Shumin Deng. 2024{\natexlab{f}}.
  \newblock \href {https://doi.org/10.18653/v1/2024.acl-long.782} {Exploring collaboration mechanisms for {LLM} agents: {A} social psychology view}.
  \newblock In \emph{Proceedings of the 62nd Annual Meeting of the Association for Computational Linguistics}, pages 14544--14607.
  
  \bibitem[{Zhang et~al.(2025{\natexlab{a}})Zhang, Liu, and Pan}]{DBLP:journals/expert/ZhangLP25}
  Qin Zhang, Ziqi Liu, and Shirui Pan. 2025{\natexlab{a}}.
  \newblock \href {https://doi.org/10.1109/MIS.2024.3517792} {The rise of small language models}.
  \newblock \emph{{IEEE} Intell. Syst.}, 40(1):30--37.
  
  \bibitem[{Zhang et~al.(2023{\natexlab{a}})Zhang, Chen, Bukharin, He, Cheng, Chen, and Zhao}]{DBLP:conf/iclr/ZhangCBH0CZ23}
  Qingru Zhang, Minshuo Chen, Alexander Bukharin, Pengcheng He, Yu~Cheng, Weizhu Chen, and Tuo Zhao. 2023{\natexlab{a}}.
  \newblock \href {https://openreview.net/forum?id=lq62uWRJjiY} {Adaptive budget allocation for parameter-efficient fine-tuning}.
  \newblock In \emph{The Eleventh International Conference on Learning Representations}. OpenReview.net.
  
  \bibitem[{Zhang et~al.(2023{\natexlab{b}})Zhang, Chen, Shen, Ding, Tenenbaum, and Gan}]{DBLP:conf/iclr/ZhangCSDTG23}
  Shun Zhang, Zhenfang Chen, Yikang Shen, Mingyu Ding, Joshua~B. Tenenbaum, and Chuang Gan. 2023{\natexlab{b}}.
  \newblock \href {https://openreview.net/forum?id=Lr8cOOtYbfL} {Planning with large language models for code generation}.
  \newblock In \emph{The Eleventh International Conference on Learning Representations}.
  
  \bibitem[{Zhang et~al.(2025{\natexlab{b}})Zhang, Zheng, Wu, Zhang, Lin, Yu, Liu, Zhou, and Lin}]{DBLP:journals/corr/abs-2501-07301}
  Zhenru Zhang, Chujie Zheng, Yangzhen Wu, Beichen Zhang, Runji Lin, Bowen Yu, Dayiheng Liu, Jingren Zhou, and Junyang Lin. 2025{\natexlab{b}}.
  \newblock \href {https://doi.org/10.48550/ARXIV.2501.07301} {The lessons of developing process reward models in mathematical reasoning}.
  \newblock \emph{CoRR}, abs/2501.07301.
  
  \bibitem[{Zhao et~al.(2023)Zhao, Zhou, Li, Tang, Wang, Hou, Min, Zhang, Zhang, Dong, Du, Yang, Chen, Chen, Jiang, Ren, Li, Tang, Liu, Liu, Nie, and Wen}]{DBLP:journals/corr/abs-2303-18223}
  Wayne~Xin Zhao, Kun Zhou, Junyi Li, Tianyi Tang, Xiaolei Wang, Yupeng Hou, Yingqian Min, Beichen Zhang, Junjie Zhang, Zican Dong, Yifan Du, Chen Yang, Yushuo Chen, Zhipeng Chen, Jinhao Jiang, Ruiyang Ren, Yifan Li, Xinyu Tang, Zikang Liu, Peiyu Liu, Jian{-}Yun Nie, and Ji{-}Rong Wen. 2023.
  \newblock \href {https://doi.org/10.48550/ARXIV.2303.18223} {A survey of large language models}.
  \newblock \emph{CoRR}, abs/2303.18223.
  
  \bibitem[{Zhao et~al.(2024)Zhao, Yin, Zeng, Wang, Shi, Lyu, Wang, Luo, and Zhang}]{DBLP:journals/corr/abs-2411-14405}
  Yu~Zhao, Huifeng Yin, Bo~Zeng, Hao Wang, Tianqi Shi, Chenyang Lyu, Longyue Wang, Weihua Luo, and Kaifu Zhang. 2024.
  \newblock \href {https://doi.org/10.48550/ARXIV.2411.14405} {Marco-o1: Towards open reasoning models for open-ended solutions}.
  \newblock \emph{CoRR}, abs/2411.14405.
  
  \bibitem[{Zhao et~al.(2022)Zhao, Jin, and Yu}]{DBLP:journals/corr/abs-2202-13876}
  Zhengyun Zhao, Qiao Jin, and Sheng Yu. 2022.
  \newblock \href {https://arxiv.org/abs/2202.13876} {Pmc-patients: {A} large-scale dataset of patient notes and relations extracted from case reports in pubmed central}.
  \newblock \emph{CoRR}, abs/2202.13876.
  
  \bibitem[{Zheng et~al.(2023)Zheng, Liu, Xie, Li, and Li}]{DBLP:journals/corr/abs-2304-09797}
  Chuanyang Zheng, Zhengying Liu, Enze Xie, Zhenguo Li, and Yu~Li. 2023.
  \newblock \href {https://doi.org/10.48550/arXiv.2304.09797} {Progressive-hint prompting improves reasoning in large language models}.
  \newblock \emph{CoRR}, abs/2304.09797.
  
  \bibitem[{Zheng et~al.(2024)Zheng, Zhang, Zhang, Ye, Luo, and Ma}]{DBLP:journals/corr/abs-2403-13372}
  Yaowei Zheng, Richong Zhang, Junhao Zhang, Yanhan Ye, Zheyan Luo, and Yongqiang Ma. 2024.
  \newblock \href {https://doi.org/10.48550/ARXIV.2403.13372} {Llamafactory: Unified efficient fine-tuning of 100+ language models}.
  \newblock \emph{CoRR}, abs/2403.13372.
  
  \bibitem[{Zhou et~al.(2024{\natexlab{a}})Zhou, Yan, Shlapentokh{-}Rothman, Wang, and Wang}]{DBLP:conf/icml/ZhouYSWW24}
  Andy Zhou, Kai Yan, Michal Shlapentokh{-}Rothman, Haohan Wang, and Yu{-}Xiong Wang. 2024{\natexlab{a}}.
  \newblock \href {https://openreview.net/forum?id=njwv9BsGHF} {Language agent tree search unifies reasoning, acting, and planning in language models}.
  \newblock In \emph{Forty-first International Conference on Machine Learning}.
  
  \bibitem[{Zhou et~al.(2024{\natexlab{b}})Zhou, Shi, Song, Yang, Jin, Guo, and Li}]{LAWGPT-zh}
  Zhi Zhou, Jiang{-}Xin Shi, Peng{-}Xiao Song, Xiaowen Yang, Yi{-}Xuan Jin, Lan{-}Zhe Guo, and Yufeng Li. 2024{\natexlab{b}}.
  \newblock \href {https://doi.org/10.48550/ARXIV.2406.04614} {Lawgpt: {A} chinese legal knowledge-enhanced large language model}.
  \newblock \emph{CoRR}, abs/2406.04614.
  
  \bibitem[{Zhou et~al.(2024{\natexlab{c}})Zhou, Hu, Zhao, Zhang, and Liu}]{DBLP:conf/ijcai/ZhouHZZL24}
  Zihao Zhou, Bin Hu, Chenyang Zhao, Pu~Zhang, and Bin Liu. 2024{\natexlab{c}}.
  \newblock \href {https://www.ijcai.org/proceedings/2024/627} {Large language model as a policy teacher for training reinforcement learning agents}.
  \newblock In \emph{Proceedings of the Thirty-Third International Joint Conference on Artificial Intelligence}, pages 5671--5679.
  
  \bibitem[{Zhuge et~al.(2024)Zhuge, Wang, Kirsch, Faccio, Khizbullin, and Schmidhuber}]{DBLP:conf/icml/ZhugeWKFKS24}
  Mingchen Zhuge, Wenyi Wang, Louis Kirsch, Francesco Faccio, Dmitrii Khizbullin, and J{\"{u}}rgen Schmidhuber. 2024.
  \newblock \href {https://openreview.net/forum?id=uTC9AFXIhg} {Gptswarm: Language agents as optimizable graphs}.
  \newblock In \emph{Forty-first International Conference on Machine Learning}.
  
  \end{thebibliography}
\end{document}